%% file: sn-article.tex
\documentclass[pdflatex,sn-nature]{sn-jnl}

\usepackage{graphicx}%
\usepackage{multirow}%
\usepackage{amsmath,amssymb,amsfonts}%
\usepackage{amsthm}%
\usepackage{mathrsfs}%
\usepackage[title]{appendix}%
\usepackage{xcolor}%
\usepackage{textcomp}%
\usepackage{manyfoot}%
\usepackage{booktabs}%
\usepackage{algorithm}%
\usepackage{algorithmicx}%
\usepackage{algpseudocode}%
\usepackage{listings}%
\usepackage{longtable}%
\usepackage{hyperref}
\usepackage{lineno}

\theoremstyle{thmstyleone}%
\theoremstyle{thmstyletwo}%

\theoremstyle{thmstylethree}%

\renewcommand{\tablename}{Supplementary Table}

\begin{document}

\title[Article Title]{Emulating Clinician Cognition via Self-Evolving Deep Clinical Research}


\author[1]{\fnm{Ruiyang} \sur{Ren}}
\equalcont{These authors contributed equally to this work.}

\author[1]{\fnm{Yuhao} \sur{Wang}}
\equalcont{These authors contributed equally to this work.}

\author[1]{\fnm{Yunsen} \sur{Liang}}

\author[2]{\fnm{Lan} \sur{Luo}}

\author*[3]{\fnm{Jing} \sur{Liu}}

\author*[3]{\fnm{Haifeng} \sur{Wang}}

\author*[4]{\fnm{Cong} \sur{Feng}}

\author[5]{\fnm{Yinan} \sur{Zhang}}

\author[5]{\fnm{Chunyan} \sur{Miao}}

\author[1]{\fnm{Ji-Rong} \sur{Wen}}

\author*[1]{\fnm{Wayne Xin} \sur{Zhao}}\email{batmanfly@ruc.edu.cn}

\affil[1]{\orgdiv{Gaoling School of Artificial Intelligence}, \orgname{Renmin University of China}, \orgaddress{\city{Beijing}, \country{China}}}

\affil[2]{\orgname{Peking University Third Hospital}, \orgaddress{\city{Beijing}, \country{China}}}

\affil[3]{\orgname{Baidu Inc.}, \orgaddress{\city{Beijing}, \country{China}}}

\affil[4]{\orgname{Chinese PLA General Hospital}, \orgaddress{\city{Beijing}, \country{China}}}

\affil[5]{\orgdiv{Joint NTU-UBC Research Centre of Excellence in Active Living for the Elderly}, \orgname{Nanyang Technological University}, \orgaddress{\country{Singapore}}}


\abstract{
Clinical diagnosis is a complex cognitive process, grounded in dynamic cue acquisition and continuous expertise accumulation. Yet most current artificial intelligence (AI) systems are misaligned with this reality—treating diagnosis as single-pass retrospective prediction while lacking auditable mechanisms for governed improvement. We developed DxEvolve, a self-evolving diagnostic agent that bridges these gaps through an interactive deep clinical research workflow. The framework autonomously requisitions examinations and continually externalizes clinical experience from increasing encounter exposure as diagnostic cognition primitives. On the MIMIC-CDM benchmark, DxEvolve improved diagnostic accuracy by 11.2\% on average over backbone models and reached 90.4\% on a reader-study subset, comparable to the clinician reference (88.8\%). DxEvolve improved accuracy on an independent external cohort by 10.2\% (categories covered by the source cohort) and 17.1\% (uncovered categories) compared to the competitive method. By transforming experience into a governable learning asset, DxEvolve supports an accountable pathway for the continual evolution of clinical AI. 
}

\maketitle

\input{secs/intro}
\input{secs/results}
\input{secs/discussion}

\input{secs/methods}

\section*{Data availability}
The MIMIC-IV dataset is available via PhysioNet subject to completion of the required data-access training and a data use agreement. The MIMIC-CDM benchmark used in this study is derived from MIMIC-IV and is available from the original release at \url{https://physionet.org/content/mimic-iv-ext-cdm} under the same terms. After obtaining access to MIMIC-CDM, the data preprocessing and cohort-splitting scripts used in this study (to reproduce the non-overlapping accrual and evaluation partitions) are available at \url{https://github.com/RUCAIBox/DxEvolve}.
The external cohort from the Chinese PLA General Hospital is not publicly available due to institutional data-governance requirements. Access to the minimum dataset necessary to reproduce the external-cohort analyses may be considered for qualified researchers, subject to approval by the hospital’s data governance procedures and execution of an appropriate data-use agreement; requests should be directed to the corresponding authors.

\section*{Code availability}
The code for DxEvolve is available at \url{https://github.com/RUCAIBox/DxEvolve}. All prompts used in DxEvolve are included in the Supplementary Information.

\nolinenumbers

\bibliography{sn-bibliography}

\clearpage
\input{secs/supplement}

\end{document}

%% file: secs/intro.tex
\section{Introduction}\label{sec1}

The mastery of diagnostic reasoning represents a defining hallmark of clinical expertise, a sophisticated cognitive process where rigorous investigation and experiential growth are inextricably linked~\cite{graber2005diagnostic, singh2015advancing, singh2014frequency, norman2017causes, dalal2025adverse}. In routine care, a seasoned clinician does not merely identify a disease from a static set of symptoms; they act as a dynamic investigator, navigating uncertainty through active, evidence-driven inquiry~\cite{ball2016improving, schwartzstein2025critical}. Moreover, each patient encounter serves as a feedback loop through which clinicians refine their internal mental scripts. Over time, these refinements accumulate into transferable experiential policies that make future decisions more robust and less prone to error~\cite{mahajan2025cognitive, ferber2025development, nenadic_physicians_2026}. {This dual capacity} for systematic investigation and continuous self-improvement underpins the maturation of clinical mastery.


Despite remarkable proficiency in medical knowledge synthesis~\cite{singhal2023large, achiam2023gpt, eriksen2024use, savage2024diagnostic, qiu2025quantifying}, current AI systems remain fundamentally misaligned with the cognitive architecture of human expertise. First, a profound process gap exists~\cite{gong2025knowledge, mccoy2025assessment, bean_reliability_2026}: most clinical AI systems treat diagnosis as a static, full-information task, collapsing the step-wise investigative rigor of the bedside into a single retrospective prediction~\cite{han2024comparative, kaczmarczyk2024evaluating, mcduff2025towards, zoller2025human, bhasuran2025preliminary, li2025macd, chen2025enhancing, zhao2026agentic}. Second and more critically, a developmental misalignment persists: whereas clinical mastery thrives on the refletive consolidation of experience, these systems function as ossified snapshots of their training data. Devoid of mechanisms to distill longitudinal practice into transferable experiences~\cite{charlin2007scripts, xu2025amem}, parameter-based updating leaves much of the learned behavior implicit. This creates a dual challenge of clinical governance: it lacks clinical auditability, as the latent logic accrued over time remains impervious to human inspection~\cite{li2024agent, us2024transparency, babic2025general, liu2025generalist}, and it precludes procedural governance, leaving the system immune to expert intervention or alignment with evolving standards~\cite{kore2024empirical, subasri2025detecting, dong2025memory}. Consequently, many systems lack an auditable, governed pathway for learning from practice—an ability that in medicine is not merely advantageous but integral to safety.

Addressing these cognitive misalignments necessitates a conceptual pivot: reconceptualizing the diagnostic process not as a mere route to a prediction, but as the essential substrate for longitudinal evolution. To faithfully emulate human diagnostic reasoning, an agent must navigate a structured investigative framework that produces traceable trajectories of evidence acquisition and hypothesis refinement that mirror the uncertainty-laden nature of clinical practice~\cite{moor2023foundation, tu2025towards, nori2025sequential}. 
Such trajectories provide the necessary learning substrate: they expose what was asked, observed and inferred at each step, enabling post hoc attribution, review and distillation of reusable experience artifacts rather than embedding all adaptation implicitly in model parameters~\cite{rajpurkar2022ai}.
By forging a symbiotic link between procedural rigor and governable evolution, it becomes possible to develop agents that not only achieve expert-level performance but also continuously cultivate their mastery that is aligned with the rigorous standards of the medical community.


In this study, we introduce DxEvolve, a self-evolving diagnostic agent that reconciles the identified gaps in existing medical AI systems by integrating a dynamic investigative workflow with an explicit experiential learning mechanism (Fig.~\ref{fig:framework}). At its foundation, DxEvolve operationalizes diagnosis through deep clinical research (DCR), an evidence-centered paradigm that reconfigures static prediction into active inquiry, synthesizing clinical findings with external medical knowledge. Within this substrate, the agent actively requisitions evidence, refines diagnostic hypotheses as cues emerge, and grounds every decision in observations with traceable provenance. Crucially, DxEvolve leverages these high-fidelity trajectories to support longitudinal self-evolution by distilling clinical encounters into diagnostic cognition primitives (DCPs)—explicit carriers of clinical experiments that link salient presentation patterns to actionable workup strategies and diagnostic insights. 
Unlike the opaque black-box updates, DCPs provide a portable repository of clinical expertise that can be selectively recalled to navigate future uncertainty. This architecture establishes a transparent pathway for clinician-led oversight and continuous improvement, while offering the practical advantage of bypassing the computationally-intensive and inflexible cycles of offline retraining.
Systematic evaluation on the MIMIC-CDM benchmark~\cite{hager_evaluation_2024} demonstrates that DxEvolve consistently enhances diverse backbone models, yielding an 11.2\% mean accuracy gain over the competitive baseline system.
Rather than relying on specific models, the framework’s efficacy is architectural: when integrated with state-of-the-art backbones, it attained expert-level proficiency under stringent dynamic constraints, achieving 90.4\% accuracy and surpassing the 88.8\% human expert (Fig.~\ref{fig:main_results}c).
Beyond static benchmarks, independent validation at the Chinese PLA General Hospital confirmed the framework's robust portability across institutional and linguistic boundaries.
The DCR architecture and distilled DCP repository yielded a 10.2\% accuracy gain on translated records and a 11.9\% improvement on raw Chinese documentation, with advantages extending to diagnostic categories entirely absent from the initial repository (17.1\% gain). 

This sustained performance is underpinned by an evolution process that resolves the developmental misalignment characteristic of static systems. We observed a longitudinal maturation effect, where experience harvested from later-stage encounters possessed higher diagnostic utility than earlier encounters. This evolution is further characterized by an error-driven dividend, where heuristics distilled from diagnostic failures catalyzed greater performance gains than those from successes. Process-level analyses confirm that DxEvolve’s investigative behavior aligns with real-world clinical practices and established clinical guidelines, ensuring that its progression is grounded in sound medical heuristics rather than statistical artifacts.

Together, these findings advance a view of clinical AI systems in which competence is defined not only by snapshot performance, but by how reliably an agent improves with exposure when diagnosis is executed as procedural evidence acquisition under workflow constraints. Our findings demonstrate that diagnostic excellence is not merely a function of static medical knowledge utilization, but a dynamic capability realized through the synergy of structured investigative workflows and progressive experiential maturation. By operationalizing these core pillars of human expertise, DxEvolve establishes that expert-level proficiency emerges when AI moves beyond statistical prediction toward the active, longitudinal cultivation of clinical wisdom. This framework provides a deployable path for clinical systems that couples workflow faithfulness with governance, supporting inspection, curation and controlled updating as standards of care and medical evidence evolve. To facilitate future research in this direction, we provide open access to our DxEvolve agentic system.

%% file: secs/results.tex
\begin{figure}
    \centering
    \includegraphics[width=1\linewidth]{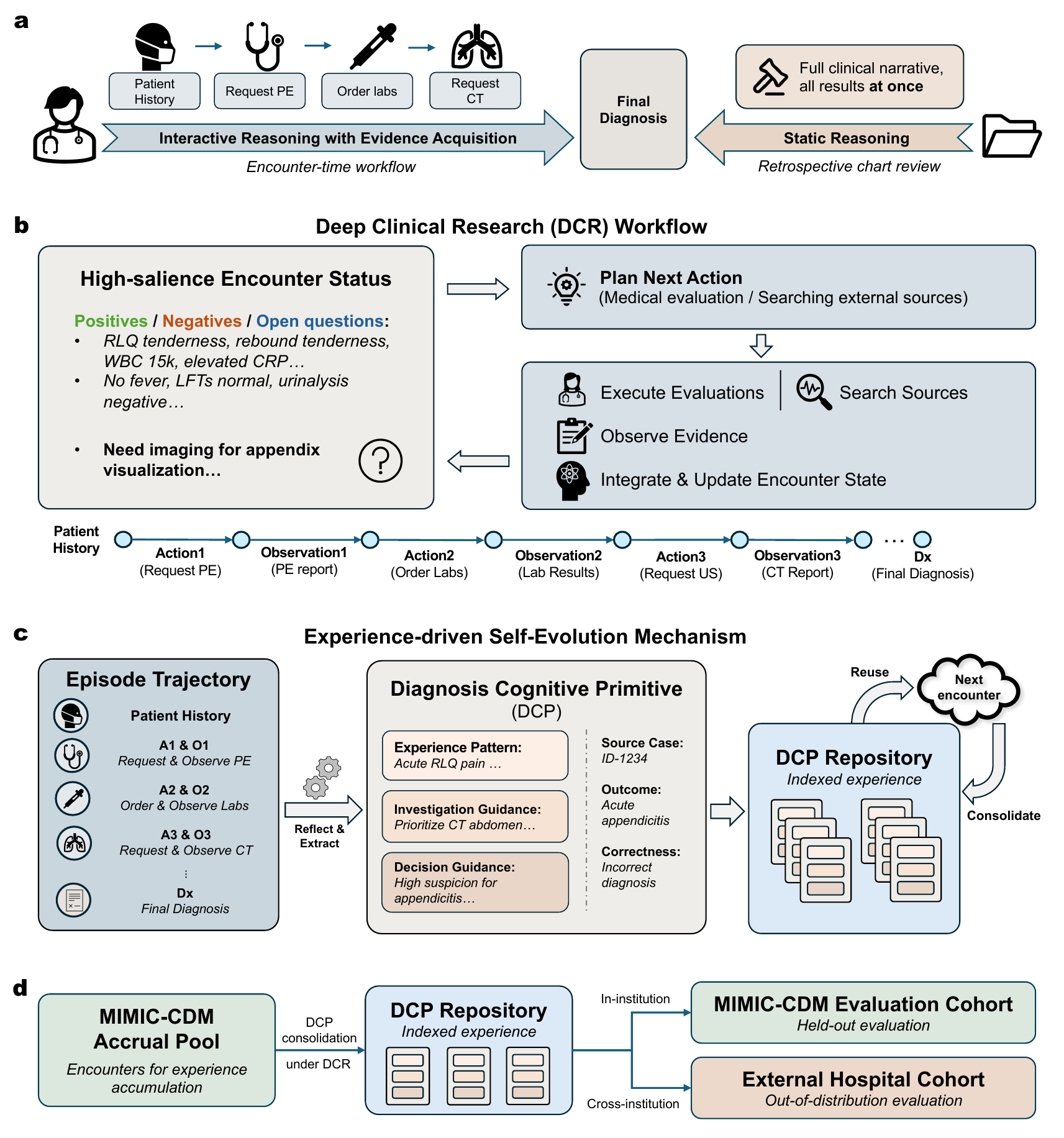}
    \caption{\textbf{DxEvolve: workflow-aligned diagnosis with experience-driven self-evolution.}
\textbf{a}, DxEvolve frames diagnosis as evidence-centered sequential reasoning, contrasting the static, single-pass inference typical of retrospective evaluations using complete records.
\textbf{b}, Deep clinical research (DCR) workflow. From the patient history context, the agent iteratively plans the next step, requests evaluations (physical examination, laboratory tests and imaging) and, when necessary, consults external sources (guidelines and PubMed); only requested observations are revealed and are integrated into a compact high-salience encounter state to guide subsequent actions until final diagnosis.
\textbf{c}, Diagnostic cognition primitives (DCPs). After each diagnosis reasoning, DxEvolve consolidates a DCP from the trajectory, consisting of a retrievable presentation pattern and evidence-linked guidance for investigation planning and diagnostic decision-making; DCPs are indexed in a repository and selectively reused in later encounters as an action like medical evaluation and searching external sources under the same DCR workflow.
\textbf{d}, Cohorts and protocol. DCPs are built from a MIMIC-CDM accrual pool that is strictly non-overlapping with evaluation encounters, then assessed on a held-out in-distribution MIMIC-CDM cohort and an external hospital cohort for out-of-distribution evaluation.}
    \label{fig:framework}
\end{figure}

\section{Results}\label{sec2}

\subsection{Experimental design and the DxEvolve framework}\label{subsec2-1}

To bridge the gap between static biomedical knowledge and dynamic clinical reasoning (Fig.~\ref{fig:framework}a), we developed DxEvolve to operationalize this dynamic reasoning process by coupling a high-fidelity investigative workflow with a mechanism for explicit experiential growth. The framework is sustained by two synergistic pillars. First, the deep clinical research (DCR) workflow ensures that every diagnostic step remains grounded in a traceable evidence base (Fig.~\ref{fig:framework}b). Second, a self-evolution mechanism distills these investigative trajectories into diagnostic cognition primitives (DCPs), effectively transforming individual patient encounters into a library of reusable, governable clinical wisdom (Fig.~\ref{fig:framework}c).

We designed an evaluation roadmap to rigorously test this framework (Fig.~\ref{fig:framework}d). First, we utilized the MIMIC-CDM benchmark~\cite{hager_evaluation_2024}, a curated dataset of 2,400 acute abdominal presentations designed specifically for stepwise diagnosis. For primary comparisons, we predefined a held-out evaluation cohort ($n$=400) randomly sampled from MIMIC-CDM and reserved all remaining non-overlapping encounters exclusively for DCP accrual; unless noted otherwise, all analyses involving DCP retrieval use this fixed accrual pool under the same split. To provide a direct anchor to human expertise, we further validated DxEvolve against another encounter split from a published clinician-benchmarked reader-study subset~\cite{hager_evaluation_2024} ($n$=80) and reserved all remaining non-overlapping encounters exclusively for DCP accrual in this setting.

Finally, to ensure the robustness extends beyond curated environments, we conducted external validation using an independent cohort from the Chinese PLA General Hospital ($N$=293). This real-world dataset, which includes diagnostic categories both overlapping with and absent from the primary benchmark, provides a stringent test of DxEvolve’s generalizability across differing healthcare systems, institutional workflows, and documentation practices. All evaluations were conducted in accordance with strict data-governance protocols, utilizing locally deployed models to ensure patient privacy and institutional compliance (``\hyperref[subsec:methods-ethics]{Ethics approval and governance}'', Methods).

\begin{figure}[htbp!]
    \centering
    \includegraphics[width=1\linewidth]{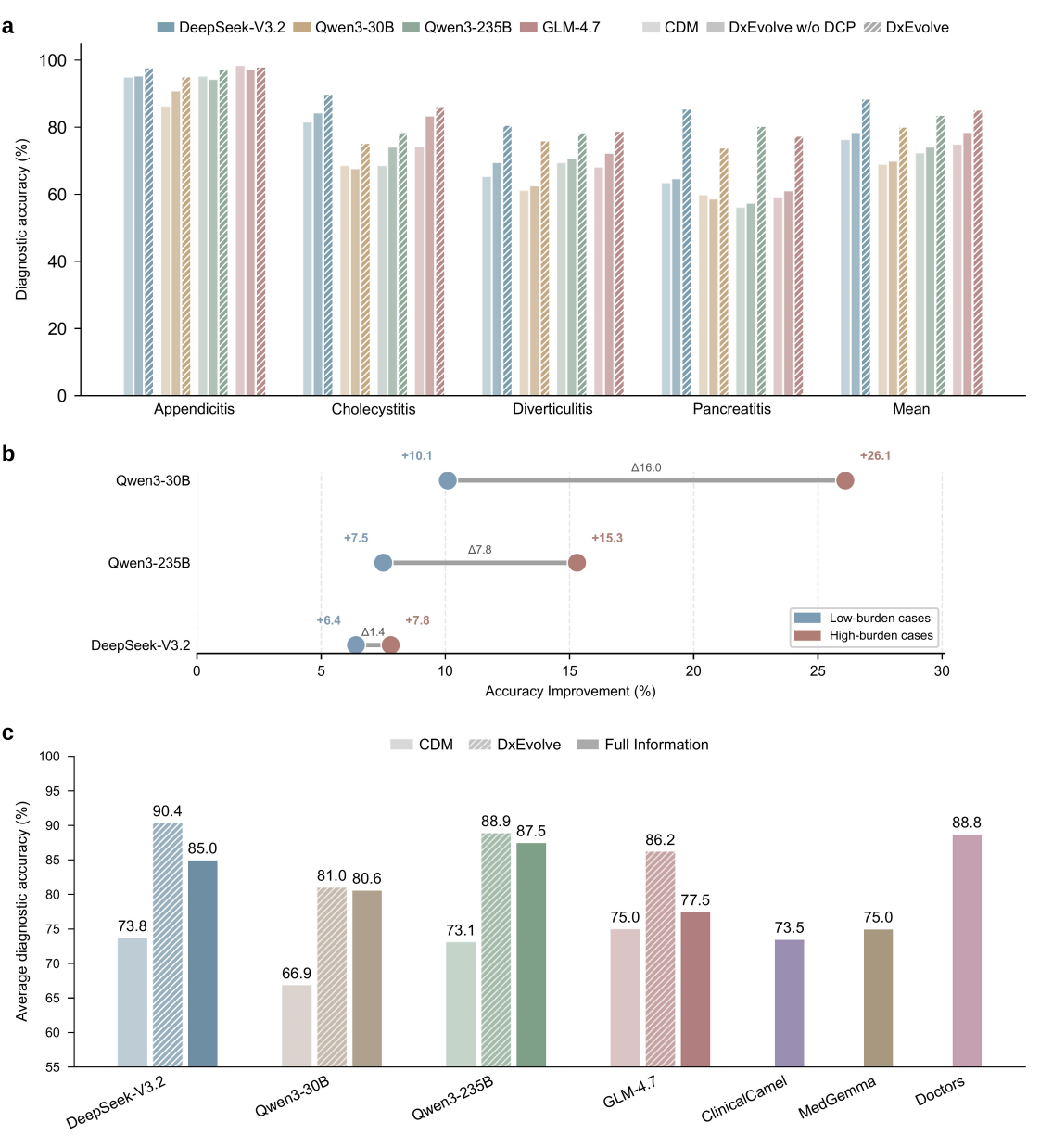}
    \caption{\textbf{Main diagnostic performance results on MIMIC-CDM.}
    \textbf{a}, Diagnosis accuracy on the MIMIC-CDM evaluation cohort ($n$=400), reported per pathology and as the average. For each base LLM (color), we compare the CDM baseline, DxEvolve without DCP retrieval (DxEvolve w/o DCP), and DxEvolve over multiple seeds.
\textbf{b}, Accuracy improvement of DxEvolve over the CDM baseline stratified by encounter-level diagnostic burden (easy versus hard). Points show the stratum-specific improvement for each base LLM; annotations indicate the improvement in each stratum and the between-stratum difference.
\textbf{c}, Diagnosis accuracy on a reader-study subset of MIMIC-CDM ($n$=80). Bars report average diagnostic accuracy for CDM and DxEvolve distinguished by light and dark shades of the same color, together with single-pass full-information (FI) inference (hatched). Specialist medical LLMs with limited action compliance are reported under FI only. The clinician reference (Doctors) corresponds to the published reader-study subset with full information available~\cite{hager_evaluation_2024}.}
    \label{fig:main_results}
\end{figure}

\subsection{DxEvolve achieves clinician-level diagnostic performance}
\label{subsec2-2}

We first evaluated DxEvolve on the MIMIC-CDM evaluation cohort ($n$=400), where Fig.~\ref{fig:main_results}a exhibited consistent diagnosis accuracy gains ($P<$0.001) across all base LLM backbones comparing with the established CDM baseline~\cite{hager_evaluation_2024} (11.2\% mean accuracy gain) and DxEvolve w/o DCP (9.1\% gain). Ablating clinical guideline and PubMed retrieval resulted in only a modest mean accuracy decrease (0.9\%), suggesting that the core gains primarily arise from workflow scaffolding and experience retrieval, with external retrieval providing complementary support in selected cases.
Critically, as these gains were achieved using off-the-shelf backbones without weight updates, the improvements reflect the efficacy of the proposed investigative workflow and experiential mechanisms rather than task-specific fine-tuning.

To characterize the utility of DxEvolve across different clinical scenarios, we stratified encounters by investigative complexity, utilizing the evidence-acquisition volume of the baseline model as a proxy for diagnostic burden. DxEvolve improved accuracy across all strata, with the most pronounced gains concentrated in the high-burden group, representing a 40\%--169\% relative increase in gain magnitude over low-burden counterparts (Fig.~\ref{fig:main_results}b). 

We next evaluate DxEvolve against human expertise using a reader-study subset of the MIMIC-CDM dataset~\cite{hager_evaluation_2024} ($n$=80). In the original reader study, clinicians issued retrospective diagnoses under a full-information (FI) regime, where all evidence was provided upfront. In contrast, DxEvolve operated under a significantly more stringent, workflow-aligned regime, requiring it to autonomously decide which evidence to acquire and when. Despite this informational disadvantage, DxEvolve attained expert-level proficiency: paired with state-of-the-art backbones, the agent achieved 90.4\% accuracy, surpassing the 88.8\% human expert (Fig.~\ref{fig:main_results}c). Notably, the clinician reference comes from the published reader-study subset under FI conditions; we use it as an anchor for human-level performance rather than a head-to-head comparison under matched information access. 
Intriguingly, DxEvolve surpassed the corresponding single-pass FI baselines across base large language models~(LLMs), including medical-domain LLMs~(ClinicalCamel and MedGemma) evaluated under the FI regime due to their inability to comply with interactive action constraints~(Fig.~\ref{fig:main_results}c). This advantage is consistent with two complementary mechanisms: first, the DCR workflow provides a reasoning scaffold that maintains clinical saliency and prevents the ``cue dilution'' common in long, unstructured records; and second, DCP-guided evolution sharpens uncertainty calibration, allowing the agent to prioritize decisive findings. 

In summary, these results demonstrate that DxEvolve couples workflow-aligned execution with longitudinal self-evolution to reach expert-level diagnostic proficiency. By externalizing improvement through explicit clinical experiences rather than opaque parametric changes, the system provides an auditable pathway for achieving high-fidelity diagnostic performance that is robust to the complexities of the real-world clinical environment.

\begin{figure}
    \centering
    \includegraphics[width=1\linewidth]{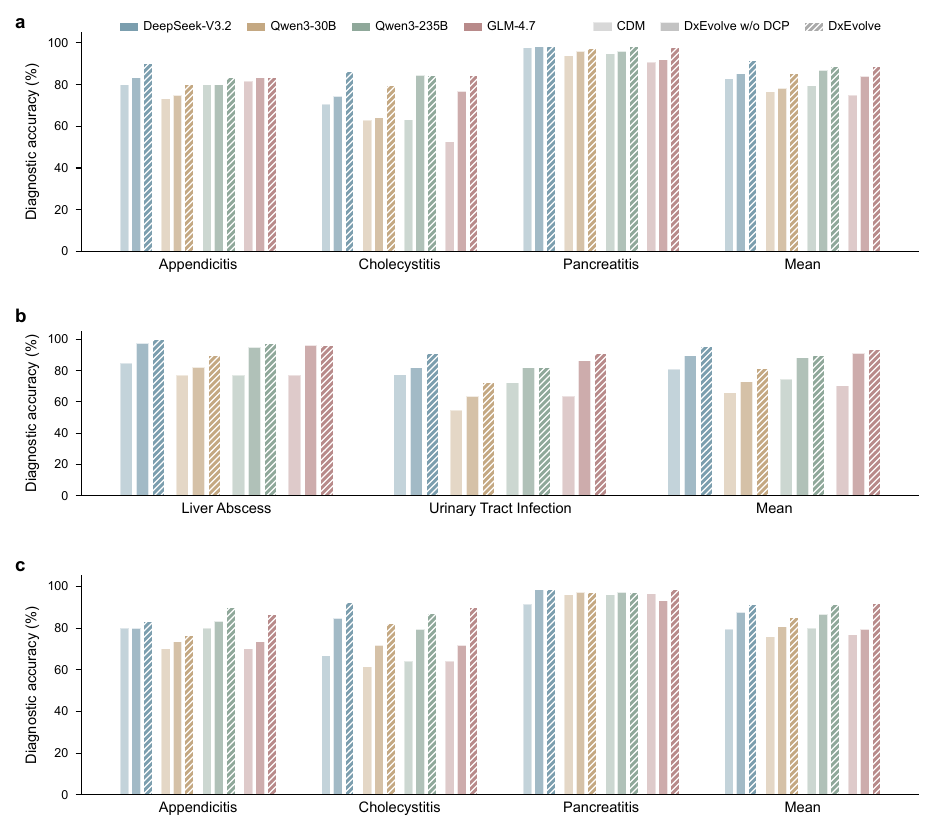}
    \caption{\textbf{External validation on an independent hospital cohort.}
    \textbf{a}, Diagnostic accuracy on diagnoses overlapping with MIMIC-CDM (appendicitis, cholecystitis and pancreatitis) and their mean, evaluated using standardized English translations of the structured records.
    \textbf{b}, Category-level transfer on diagnoses that were never used for DCP accrual (liver abscess, urinary tract infection) and their mean, evaluated under the same protocol.
    \textbf{c}, Robustness to documentation with native institutional language, evaluated on the same external encounters using the original Chinese records.}
    \label{fig:ext_valid}
\end{figure}

\subsection{External validation supports cross-institution portability of experiential gains}\label{subsec2-3}

To evaluate the external validity of DxEvolve, we conducted independent validation on a cohort from the Chinese PLA General Hospital, representing a substantial shift (``\hyperref[subsec:methods-cohorts]{Evaluation cohorts}'' in Methods). To decouple institutional variance from linguistic factors, we applied the DCP repository distilled from 2,000 MIMIC-CDM encounters to standardized English translations of these clinical records. DxEvolve consistently elevated performance across all base LLMs, yielding a 10.2\% mean accuracy gain over the CDM baseline and a 5\% improvement over the DCP-free ablation (Fig.~\ref{fig:ext_valid}a). This sustained efficacy across distinct national and institutional contexts suggests that distilled DCPs capture trans-institutional diagnostic heuristics rather than narrow, dataset-specific shortcuts tied to the originating environment.

While overall accuracy on the external cohort was comparable to that on MIMIC-CDM, we observed notable heterogeneity across disease states. Accuracy for appendicitis and cholecystitis decreased, whereas performance on pancreatitis encounters improved. While the source of this variance likely reflects institution-specific workup pathways and documentation nuances, highlighting the necessity of evaluating clinical agents across diverse practice environments where diagnostic thresholds and recording standards may differ.

We further probed the framework's adaptability on diagnostic categories absent from the initial repository, including liver abscess and urinary tract infection (UTI). In these out-of-distribution settings, DxEvolve yielded a 17.1\% mean accuracy gain averaged across liver abscess and UTI cohorts over the CDM baseline and a 4.5\% improvement over the DCP-free ablation (Fig.~\ref{fig:ext_valid}b). Notably, while liver abscess shares the abdominal domain of the original benchmark, UTI represents a distinct clinical entity. These gains indicate that distilled DCPs encode portable, domain-agnostic heuristics that transcend specific disease labels. While the full scope of transferability across heterogeneous syndromes warrants further investigation, these results demonstrate the robust scalability of experience-guided evolution in previously unencountered clinical domains.

Finally, we assessed the cross-lingual robustness of DxEvolve by evaluating its performance on original Chinese clinical records. In this practical deployment scenario, patient encounters were processed in their native language, while the underlying reasoning framework and the accumulated DCP repository remained in English. Despite this linguistic mismatch, DxEvolve yielded an 11.9\% mean accuracy gain over the CDM baseline and a 6.3\% improvement over the DCP-free ablation (Fig.~\ref{fig:ext_valid}c). Notably, absolute diagnostic accuracy remained comparable to that achieved using standardized English translations. These observations demonstrate that the DCR framework and experiential heuristics within DxEvolve are language-agnostic, confirming the framework's viability in diverse, multilingual clinical environments.

Together, these external evaluations demonstrate that DxEvolve’s self-evolution mechanism confers substantial portability across institutional boundaries, documentation languages, and diagnostic categories. By externalizing clinical wisdom as symbolic, governable assets, the framework provides a rigorous trajectory for maintaining high-fidelity performance amidst the inherent heterogeneity of real-world clinical practice.

\begin{figure}
    \centering
    \includegraphics[width=1\linewidth]{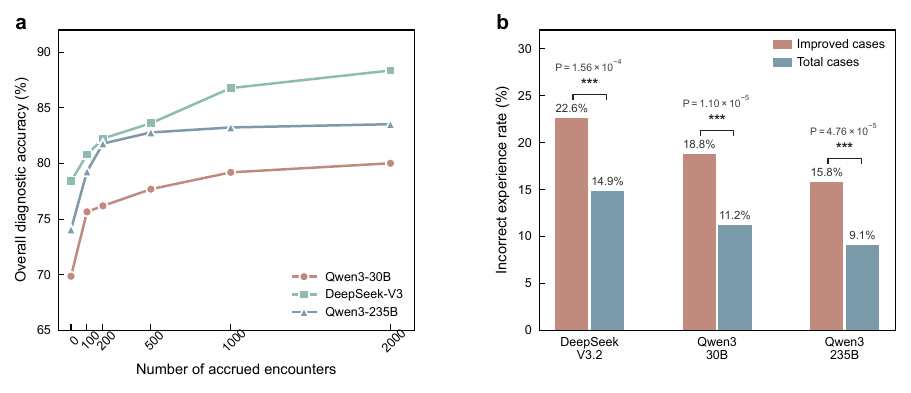}
    \caption{\textbf{Exposure-dependent self-evolution and provenance of retrieved experience.}
    \textbf{a}, Overall diagnosis accuracy on the fixed MIMIC-CDM evaluation cohort ($n$=400) as the DCP accrual pool increases, shown for three representative base LLM backbones. Accuracy improves with additional accrual encounters and then tapers, yielding a saturating learning curve.
    \textbf{b}, Provenance of retrieved experience during evaluation. Bars show the fraction of retrieved DCPs whose source accrual episode ended in an incorrect diagnosis (``incorrect experience rate''), computed separately for improvement cases and for all evaluation encounters pooled. $P$ values indicate enrichment of incorrect-source DCPs among retrievals in improvement cases.}
    \label{fig:self_evolving}
\end{figure}

\subsection{Self-evolution shows exposure-dependent scaling behavior and error-driven correction}\label{subsec2-4}


We next studied whether DxEvolve exhibits exposure-dependent improvement consistent with clinician-like development, and whether the gains can be traced to reusable experience rather than incidental trajectory variation. We therefore quantified self-evolution by scaling the pool of encounters available for DCP accrual while holding the evaluation cohort fixed (``\hyperref[subsec:methods-analysis]{Evaluation and analysis}'' in Methods).

Accuracy matured longitudinally, yielding reproducible learning curves across all evaluation schedules~(Fig.~\ref{fig:self_evolving}a), with a mean accuracy gain of 8.97\% after accrual over the first 0–1,000 encounters and a further 0.9\% gain over 1,000–2,000 encounters. While initial gains were remarkable, trajectories eventually diverged by model capacity: whereas weaker backbones reached an asymptotic plateau, more capable models sustained incremental growth throughout the accrual period. This divergence suggests that the saturation point of experience-guided evolution is governed by the base LLM's reasoning capability; stronger architectures demonstrate a superior ability to mine from complex, long-tail scenarios, effectively raising the ceiling of attainable diagnostic expertise.


To identify which experiences drive error correction, we analyzed improvement cases—encounters where DxEvolve succeeded but its baseline failed. In these cases, retrieved DCPs were significantly enriched with experiences distilled from prior diagnostic failures compared to the general retrieval distribution (Fig.~\ref{fig:self_evolving}b). This highlights an error-driven dividend, where heuristics rooted in past mistakes contribute more to subsequent performance gains. These results suggest that failures represent high-value learning events, providing the critical corrective logic necessary to navigate complex diagnostic pitfalls that successful encounters may overlook.

Together, these analyses connect exposure-dependent performance gains to an inspectable mechanism: improvement scales with accumulated experience, and the experience invoked when errors are corrected exhibits a systematic provenance structure. This motivates examining not only how the repository grows, but how the content of accrued DCPs matures with continued exposure.

\begin{figure}
    \centering
    \includegraphics[width=1\linewidth]{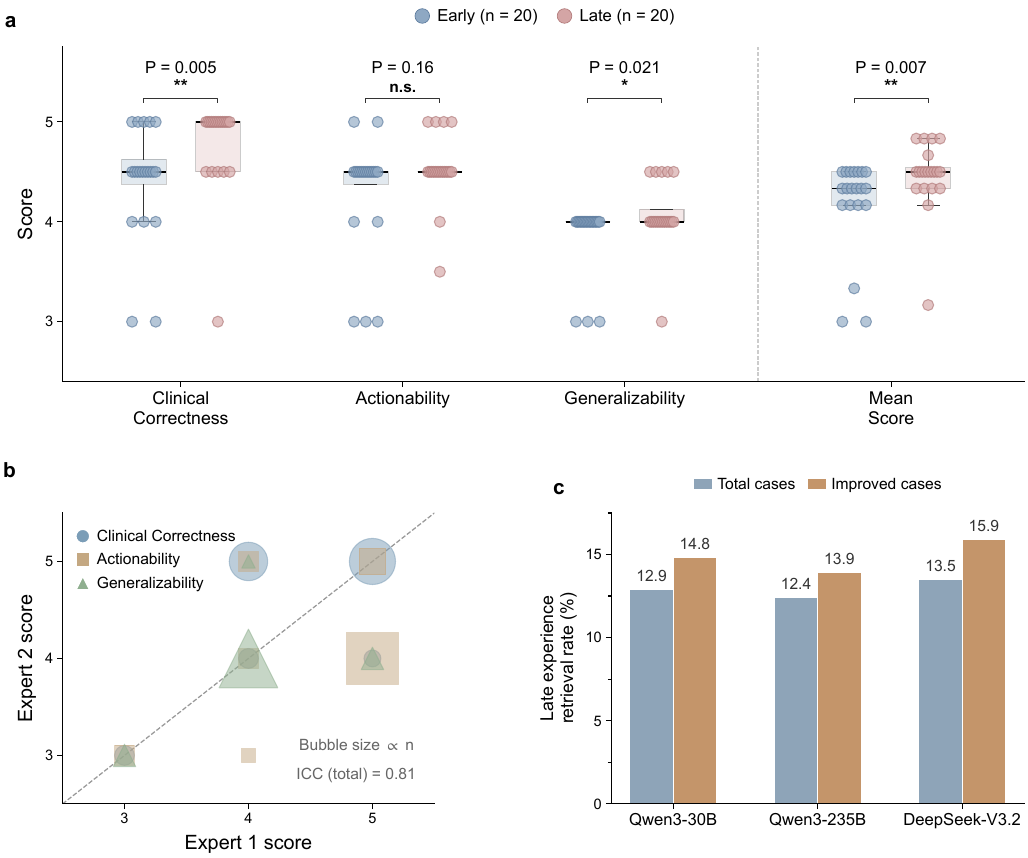}
    \caption{\textbf{Maturation of accrued experience artifacts with encounter exposure.}
    \textbf{a}, Blinded clinician ratings of diagnostic cognition primitives (DCPs) sampled from an early exposure window (encounters 1–300; $n$=20) and a late window (encounters 1700–2000; $n$=20). DCPs were scored for clinical correctness, actionability and generalizability, with the mean shown as an aggregate. Boxes denote interquartile range, centre line the median, and points individual DCPs; two-sided P values are shown (n.s., not significant).
    \textbf{b}, Inter-rater reliability of clinician ratings for the aggregate DCP score (ICC=0.81), supporting the reliability of the clinician assessment.
    \textbf{c}, Evaluation-time retrieval signal for late-stage DCPs, quantified as the fraction of retrieval events that involve DCPs in the late encounter window.}
    \label{fig:exp_quality}
\end{figure}

\subsection{Self-evolution is accompanied by progressive maturation of experience}\label{subsec2-5}

To quantify the functional maturation of the experience repository, we examined whether DCPs accrued in later developmental stages exhibit superior clinical utility and broader applicability than early-stage heuristics. This progression was validated through blinded expert assessment and comprehensive retrieval-log analyses (``\hyperref[subsec:methods-analysis]{Evaluation and analysis}'' in Methods).

In a clinician reader study blinded to study condition, we randomly sampled 20 DCPs from an early exposure window (encounters 1--300) and 20 from a late window (encounters 1700--2000). Two clinicians rated each DCP on clinical correctness (including safety concerns), actionability (guiding evidence acquisition and hypothesis refinement) and generality (reusability beyond the source encounter and pathology). The robustness of the expert evaluation framework was confirmed by high inter-rater reliability for the aggregate DCP scores (intraclass correlation coefficient (ICC)=0.81; Fig.~\ref{fig:exp_quality}b). Late-stage DCPs scored higher across dimensions than early-stage DCPs, with mean clinician rating 4.47 vs 4.17 on a 5-point scale (Fig.~\ref{fig:exp_quality}a). Both sets often contained clinically reasonable guidance, but later DCPs more consistently articulated it in reusable, action-oriented terms (for example, clearer conditional checks and escalation cues), whereas early DCPs more often remained context-bound, supporting gradual maturation with exposure.

To complement clinician ratings with a usage-based signal, we analyzed evaluation-time DCP retrieval logs. Using the same early and late exposure windows, we quantified for each DCP (i) \emph{retrieval breadth} (the number of distinct evaluation encounters in which it was retrieved) and (ii) \emph{association with error-correcting episodes} (retrieval events in encounters where DxEvolve was correct but DxEvolve w/o DCP was incorrect). 
Retrieval log analyses confirmed that late-stage DCPs possess superior functional utility. While these artifacts maintained a baseline retrieval rate of 12.4\%--13.5\% across total encounters, their prevalence increased to 13.9\%--15.9\% within error-correcting episodes (Fig.~\ref{fig:exp_quality}c). This enrichment was most pronounced in DeepSeek-V3.2.

Taken together, clinician-blinded ratings and usage-based signals converge on a consistent picture: with continued encounter exposure, DCPs become more reliably actionable and more broadly reusable, and their retrieval is increasingly enriched in error-correcting episodes. These findings support that self-evolution involves qualitative refinement of accrued experience artifacts, rather than simply expanding the size of the DCP repository.

\begin{figure}
    \centering
    \includegraphics[width=1\linewidth]{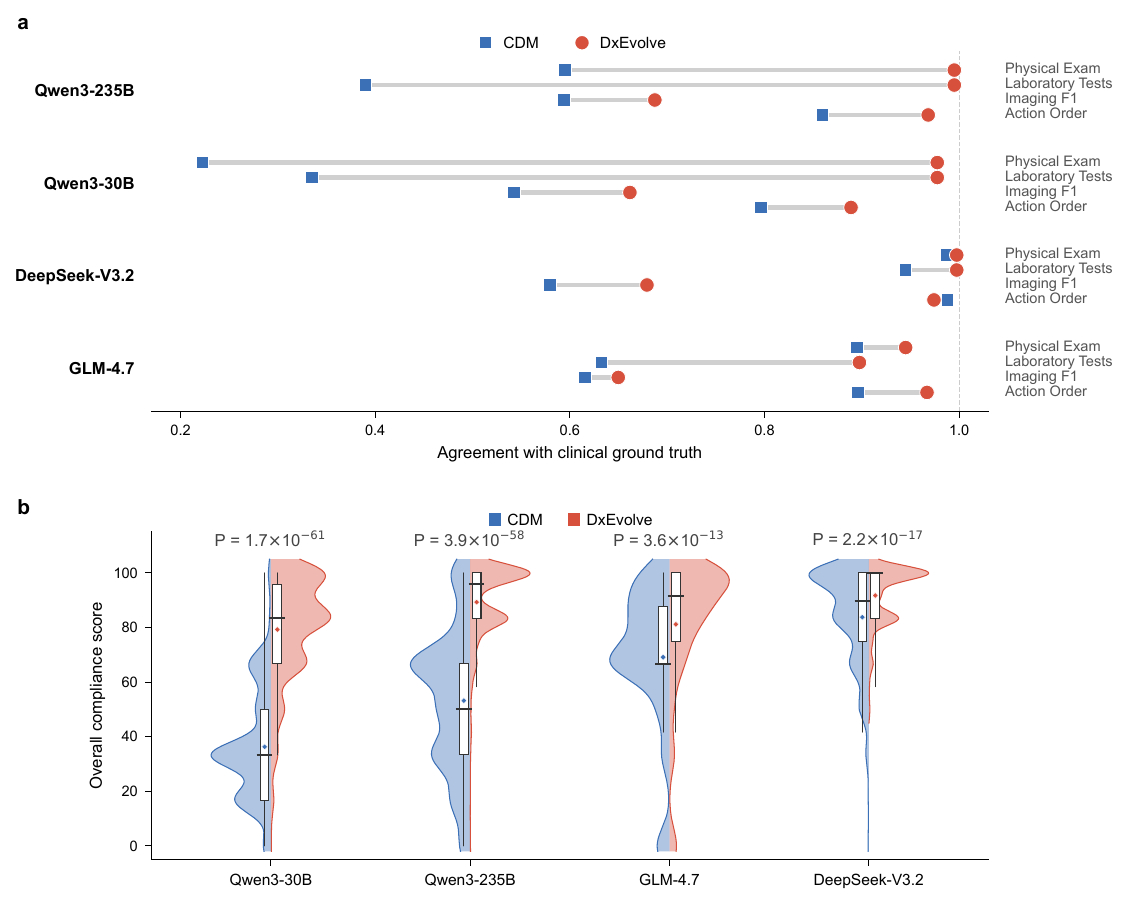}
    \caption{\textbf{DxEvolve produces more workflow-consistent investigations and shows improved alignment with clinical guidelines.}
    \textbf{a}, Workup consistency. Across the MIMIC-CDM evaluation cohort ($n$=400), DxEvolve shows higher agreement with the documented investigation trace than the standard decision-making baseline CDM for each backbone, spanning whether a physical examination was performed, overlap with recorded laboratory testing, overlap with recorded imaging (modality and region), and concordance of the investigation ordering. Points are model-level means; grey lines connect paired results for DxEvolve versus CDM under the same backbone.
    \textbf{b}, Guideline adherence. Distributions of encounter-level guideline-compliance scores, derived from the mean adherence across three dimensions: physical examination, laboratory investigations, and imaging. Violin plots show score densities; embedded boxplots indicate the median and interquartile range; points mark the mean. $P$ values are from paired two-sided comparisons.}
    \label{fig:action}
\end{figure}

\subsection{DxEvolve's evidence acquisition aligns with clinical workflows and clinical guidelines}\label{subsec2-6}

In workflow-aligned diagnosis, performance depends not only on the final diagnosis but also on whether requested investigations resemble routine care. We therefore assessed DxEvolve’s evidence-acquisition behaviour at the encounter level, measuring alignment with documented investigations and compatibility with common pathways (``\hyperref[subsec:methods-analysis]{Evaluation and analysis}'' in Methods).


Across the MIMIC-CDM evaluation cohort, DxEvolve exhibited higher consistency with recorded workups on all four trajectory-consistency measures than the standard workflow-aligned baseline (mean overall consistency across base LLMs, 0.89 and 0.68, respectively), including physical-examination execution, laboratory-test set F1, imaging (modality, region) set F1 and action-order concordance. The results indicate more reliable coverage of key investigation types and a workup sequence closer to the recorded workflow (Fig.~\ref{fig:action}a).

We further assessed workup behavior against established clinical guidelines using a conservative, three-component compliance score that captures (i) whether physical examination was performed before downstream testing, (ii) coverage of guideline-recommended laboratory categories and (iii) whether the first imaging study matched guideline-supported modality–region choices for each condition. DxEvolve achieved higher overall compliance than CDM across all evaluated backbones, with distributions shifted toward higher scores and statistically significant paired differences as shown in Fig.~\ref{fig:action}b. 

Together, these analyses indicate that DxEvolve’s improvements extend beyond end-point accuracy to more clinically compatible evidence acquisition, rather than arising from opportunistic or idiosyncratic request patterns.

%% file: secs/discussion.tex
\section{Discussion}\label{sec3}

This study presents DxEvolve, a self-evolving diagnostic agent that instantiates diagnosis as an interactive deep clinical research (DCR) workflow, in which clinical evidence is acquired procedurally through explicit evaluation actions, with optional consultation of external medical sources. DxEvolve is designed as a governed learning system over encounter-level diagnostic trajectories, supporting longitudinal self-evolution by accruing and retrieving diagnostic cognition primitives (DCPs) as reusable experience artifacts. Across a public, de-identified benchmark of clinical encounters formatted for procedural evidence acquisition, DxEvolve reaches clinician-comparable performance under interactive diagnosis. Importantly, evaluation on an external cohort from a Chinese tertiary hospital operating in a distinct healthcare system shows consistent DCP-enabled gains, supporting the portability of experience under cross-institutional shift. These findings show that workflow-aligned diagnostic agents can reach clinician-benchmarked performance while preserving auditability, reframing progress from static full-record prediction to governed, evidence-tethered execution and improvement as clinical expertise accrues.

A central contribution of DxEvolve lies in the experience-driven self-evolution mechanism, which renders encounter exposure an explicit learning signal within a workflow-aligned diagnostic process. Unlike paradigms that treat each case as a static, full-record input, where all documented findings are provided upfront, DxEvolve operates through procedural evidence acquisition and iterative hypothesis refinement under the DCR framework. This design more closely mirrors the temporal and inferential structure of routine diagnostic workups. By generating standardized, clinically auditable trajectories with explicit provenance, DxEvolve learns from practice in a manner analogous to human clinicians. 
Through this process, DCPs are accumulated into a reusable experience repository and can be retrieved to steer subsequent evidence gathering and diagnostic refinement without parameter updates. When external medical sources are consulted, their evidence can provide additional authoritative corroboration.
Empirically, diagnostic performance improved with cumulative encounter exposure, yielding a reproducible, exposure-dependent scaling curve. Notably, DCPs originating from prior diagnostic failures were enriched in improvement cases, suggesting an error-driven learning mechanism: unsuccessful episodes preferentially yield corrective effects that reduce the likelihood of repeating similar mistakes in similar clinical contexts. Because DCP-based self-evolution remains non-parametric and traceable, these primitives can be inspected, curated, or even retracted as needed. This offers a practical pathway for governed, longitudinal adaptation, a capability difficult to achieve through conventional model training.

External validation at the Chinese PLA General Hospital confirms that DxEvolve’s advantages transcend institutional boundaries, linguistic variations, and diagnostic categories. The DxEvolve’s sustained performance across translated and native Chinese documentation suggests that its distilled experiences capture portable, workflow-level logic rather than language-specific artifacts. Notably, the observed gains in diagnostic categories absent from the initial repository underscore a cross-disease generalizability essential for real-world deployment. 
Collectively, the DCR workflow provides a portable execution substrate for stepwise evidence acquisition under heterogeneous documentation, and DCP-based self-evolution supplies a reviewable mechanism for adaptation as institutions, languages and workup patterns drift. They offer a practical route to maintaining dependable diagnostic performance beyond the originating benchmark.


Beyond exposure-dependent performance gains, our results suggest that self-evolution is accompanied by a progressive improvement in the quality of accrued DCPs, echoing how clinicians’ experiential knowledge can mature with seniority rather than remaining isolated reflections. In clinician-blinded assessments, experiences accumulated later scored higher on clinical correctness, actionability and generality than earlier experiences, although both stages were broadly clinically reasonable. Consistent with this, usage-based analyses showed that later experiences were retrieved across a wider range of evaluation encounters and were more often observed in error-correcting episodes under identical workflow constraints. Together, these signals support a maturation process in which accrued experience becomes more reliably actionable and more broadly reusable, rather than simply expanding in volume. In practice, the gains from self-evolution reflect experience refinement as well as accumulation.

For workflow-aligned clinical agents, terminal diagnostic accuracy is an incomplete endpoint because the agent determines the sequence and intensity of evidence acquisition, with downstream implications for test utilization and imaging escalation. DxEvolve’s requested investigations matched encounter-recorded workups more closely than the baseline across behavioural concordance measures, and more often selected guideline-supported first-line imaging. Together with the accuracy gains, these process-level improvements suggest that the gains are not primarily explained by indiscriminate escalation of investigations. Such process alignment provides an auditable substrate for governance, enabling calibration of investigation intensity and targeted review of recurrent failure patterns.

Notwithstanding these advances, several limitations and corresponding priorities for future work warrant consideration. First, our experiments use de-identified EHR-derived records to enable reproducible, auditable measurement of evidence acquisition and experience reuse; extending this framework to prospective settings will benefit from incorporating additional real-world factors, such as clinician–patient interaction. 
Second, we observe consistent gains when applying distilled experiences to diagnostic categories beyond those represented in the initial repository, supporting portability across disease settings; broader evaluations across diverse case-mix and clinical contexts will further delineate generalizability in complex practice. Third, our current action schema emphasizes the core diagnostic-relevant actions required for diagnosis in an interactive workup setting; the framework is naturally extensible to richer actions as needed for specific clinical deployments.
These considerations motivate three next steps: (i) prospective clinician-in-the-loop studies that evaluate workflow fidelity, efficiency and patient-relevant endpoints; (ii) expanded multi-institutional and multi-specialty evaluation to characterize when and where experience-guided self-evolution generalizes; and (iii) extension of the action space to incorporate richer operational actions while preserving auditability and benchmarking comparability.

In summary, DxEvolve links workflow-aligned diagnostic investigation with longitudinal, governed improvement through experience-driven self-evolution. By operationalizing diagnosis as procedural evidence acquisition alongside auditable experience consolidation, the framework reflects two core elements of clinical expertise: systematic investigation within a patient encounter and progressive learning across a career. Consistent with this, DxEvolve reaches clinician-level performance under evaluations that emulate clinically realistic diagnostic constraints, demonstrating that sophisticated diagnostic reasoning emerges when structured investigative protocols are refined by an ever-maturing repository of DCPs. 
By externalizing learning into inspectable artifacts rather than opaque parameter updates, DxEvolve aligns AI advancement with the transparency standards essential to clinical safety. More broadly, our findings support governed, auditable self-evolution as a promising direction for clinical AI that must remain reliable as evidence and standards of care evolve.

%% file: secs/methods.tex
\section{Methods}\label{sec:methods}

\subsection{DxEvolve framework}\label{subsec:methods-DxEvolve}

DxEvolve is a self-evolving diagnostic agent that closes two coupled gaps observed in clinical AI diagnosis: a process gap between static full-information prediction and workflow-aligned stepwise evidence acquisition, and a learning gap in which apparent competence does not accumulate into more reliable evidence-consistent reasoning under uncertainty. DxEvolve operationalizes diagnosis as an evidence-centric deep clinical research workflow and the proposed self-evolution mechanism externalizes longitudinal improvement as auditable diagnostic cognition primitives, distilled from and reinvoked within the same diagnostic trajectories, without any parameter updates to the base large language models (LLMs).


At the core of each clinical encounter, DxEvolve implements a deep clinical research (DCR) framework—an agentic research protocol that treats diagnosis as evidence-driven investigation rather than single-pass prediction, while enforcing workflow-aligned constraints on evidence acquisition. Each encounter starts from the presenting complaint with limited initial context, mirroring early-stage clinical uncertainty. The agent then iteratively plans the next information need, executes a concrete acquisition action, and updates an explicit encounter state that integrates newly revealed findings with the evolving hypothesis set and a structured plan for subsequent steps. The DCR workflow thus proceeds through repeated cycles of (i) formulating the next evidence-seeking objective conditioned on the current state, (ii) acquiring the selected information through tool-mediated actions, and (iii) synthesizing the new evidence into the state to refine hypotheses and commit to the next investigative decision.

The action space is aligned with routine workup operations and includes requests for physical examination findings, laboratory testing results and imaging reports. Because evidence availability and recommended workup choices are often guided by evolving clinical guidance and best practices, relying solely on parametric model knowledge can be insufficient, particularly early in an encounter when patient-specific evidence is sparse. DxEvolve can therefore optionally invoke external medical evidence interfaces (PubMed and clinical guidelines) within the same workflow to support evidence-grounded decision-making and to reduce reliance on unsupported rationales.
Specifically, clinical guidelines are accessed via dense retrieval through semantic vector-space indexing to identify contextually relevant standards, while peer-reviewed evidence is sourced through queries to the official PubMed search utilities.

The DCR workflow can rapidly obtain long and heterogeneous text (for example, multi-parameter laboratory outputs, narrative imaging reports and retrieved documents), in which weakly relevant or incidental content may dilute clinically decisive signals. To mitigate this, DxEvolve applies context engineering by prioritizing clinically salient findings and suppressing incidental content in the running context, performing an automatic summarization step that extracts and carries forward diagnostically relevant information when needed. This mechanism preserves continuity of the diagnostic trajectory while maintaining a stable, high-signal representation to inform subsequent decisions. 
Importantly, the DCR-generated diagnostic trajectories can drive longitudinal learning with real encounter-derived workups and outcomes rather than by abstract, simulator-specific feedback.

The central innovation of DxEvolve is the longitudinal self-evolution mechanism that enables progressive improvement with clinical exposure by accumulating and reusing experience from prior episodes, without any parameter updates to the underlying base LLM. This design is motivated by clinician cognition: expertise is not only the recall of medical facts, but the ability to recognize recurring clinical patterns, anticipate high-yield investigations and apply context-appropriate decision rules shaped by prior successes and failures. This design externalizes learning into accountable experience artifacts that clinicians can audit, revise or remove, rather than relying on latent behavioural drift.

After each completed diagnostic episode in the accumulation pool, DxEvolve performs a structured post-hoc consolidation step over the trajectory and distills a diagnostic cognition primitive (DCP) optimized for reuse under uncertainty. Each DCP contains three components: experience pattern, test-ordering experience, and diagnostic decision experience. The experience pattern provides a high-salience signature for retrieval, summarizing the presentation and discriminative cues at a level intended to generalize beyond the originating patient. The test-ordering experience encodes actionable workup guidance for the stepwise setting, including high-yield next-step evaluations, contingency options when findings are equivocal and safety-oriented guardrails that reduce common omissions or inappropriate escalation. The diagnostic decision experience captures evidence-linked implications for hypothesis refinement and final decision-making, including discriminative patterns that support or refute leading hypotheses, red-flag checks, and corrective lessons when the source trajectory exposed an error mode. DCPs are written as portable guidance rather than narrative rationales.

To support mechanistic analyses and traceable governance, each DCP is stored with lightweight provenance metadata for in-depth analysis, including its exposure index, diagnostic category and whether the source episode produced a correct primary diagnosis. This provenance enables analyses of how DCP sources relate to subsequent performance gains and error correction.

During diagnosis on encounters, DxEvolve treats the DCP repository as a growing long-term memory. At the step of deciding to retrieve prior experience, the agent derives a retrieval query from its current evidence-grounded state and retrieves a small set of candidate DCPs whose experience patterns best match the current presentation. Retrieved DCPs are injected as a bounded context and applied as conditional guidance: they may steer evidence seeking, highlight discriminative cues to verify or provide evidence-linked guidance for final diagnostic commitment. To mitigate spurious memory-driven bias, DxEvolve is instructed to use a DCP only when it is compatible with the patient-specific evidence acquired so far and to disregard DCP guidance that is irrelevant with observed findings.

By combining workflow-aligned trajectories with structured DCP consolidation and evidence-compatible reuse, DxEvolve provides an accountable pathway for exposure-dependent improvement while preserving transparency and avoiding fine-tuning-induced shifts in base-model behaviour. Diagnostic reasoning trajectories and DCP examples are shown in Supplementary Section C and D.

\subsection{Data sources}\label{subsec:methods-datasources}
Benchmark experiments used MIMIC-CDM~\cite{hager_evaluation_2024}, a clinical decision-making benchmark curated from MIMIC-IV~\cite{johnson2023mimic}. MIMIC-IV is a large, de-identified electronic health record resource sourced from routine clinical care at Beth Israel Deaconess Medical Center (Boston, MA, USA), including longitudinal structured variables, laboratory measurements and linked clinical documentation~\cite{johnson2023mimic}. MIMIC-CDM inherits this real-world provenance and comprises 2{,}400 de-identified patient presentations of acute abdominal pain spanning four diagnostic categories (appendicitis, cholecystitis, diverticulitis and pancreatitis), formatted for workflow-aligned diagnosis in which additional evidence (such as physical examination findings, laboratory results and imaging reports) is revealed only when explicitly requested through the corresponding action~\cite{hager_evaluation_2024}. 

To prevent label leakage, agent-facing inputs excluded any diagnosis fields or label-bearing metadata. Evidence items were provided as structured text fields in the dataset release, with field boundaries preserved to avoid inadvertent information disclosure through formatting, concatenation or re-ordering. When multiple items of the same evidence type were available, they were retained in their original record order and were exposed only after the agent issued the matching request action.

\subsection{Evaluation cohorts}\label{subsec:methods-cohorts}
Across all experiments, we enforced strict non-overlap between encounters used for longitudinal experience accumulation (\textit{i.e.}, construction of the diagnostic cognition primitive repository, DCP) and those used for evaluation, implemented at the encounter level using unique identifiers. For primary comparisons under the deep clinical research~(DCR) workflow, we predefined a held-out MIMIC-CDM evaluation cohort of 400 encounters and kept it fixed across base models, ablations and random seeds; all remaining non-overlapping MIMIC-CDM encounters were used exclusively for DCP accrual. 

To contextualize against published clinician benchmarking, we additionally evaluated on the reader-study subset from Hager et al.\ (80 encounters; 20 per pathology)~\cite{hager_evaluation_2024}, which was treated as an independent evaluation cohort and strictly excluded from DCP accrual. On this subset, we report both workflow-aligned evaluation and single-pass full-information (FI) inference using identical underlying encounter content, differing only in the information-availability interface (complete record provided upfront for FI, with evidence-request actions disabled).

For external validation, we assembled an independent cohort of de-identified encounters (2020--2024) from the Chinese PLA General Hospital ($N$=293) curated with a standardized record structure, including appendicitis ($n$=30), cholecystitis ($n$=39) and pancreatitis ($n$=174), which match diagnostic categories in MIMIC-CDM, as well as liver abscess ($n$=39) and urinary tract infection ($n$=11). The cohort composition reflects the natural prevalence and clinical distribution of these conditions within the institution’s stream, preserving the ecological validity of the dataset and ensuring that the evaluation mirrors the diagnostic challenges encountered in unconstrained real-world practice.
All external encounters were used exclusively for out-of-distribution testing and were never used for DCP accrual. For external-cohort experiments, the DCP repository was built solely from the MIMIC-CDM accrual pool using the same base LLM as in the corresponding evaluation.

Records were harmonized to follow the MIMIC-CDM task format, preserving the initial presenting complaint and a pool of candidate evidence items retrievable through explicit requests. Imaging evidence followed the MIMIC-CDM convention by providing only the final narrative report text. Owing to source-format constraints, laboratory testing was returned as a consolidated results field, analogous to physical examination returns.

To enable controlled cross-institutional evaluation with English-prompted base models, we produced standardized English translations of the structured records using an offline, locally run translation tool with human verification. Translation was performed at the field level to preserve section boundaries and avoid reordering or merging across fields; numerical values, units and unambiguous medical abbreviations were retained.

For cross-language robustness, we additionally evaluated DxEvolve on the original Chinese structured records under the same workflow and action schema. In this setting, only the patient-specific encounter content was in Chinese, whereas prompts and the DCP repository remained in English.

\subsection{Ethics approval and governance}\label{subsec:methods-ethics}
MIMIC-IV and the derived MIMIC-CDM cohort contain de-identified patient data and were accessed via PhysioNet under the required credentialing and data-use agreements, in accordance with the dataset governance policies~\cite{johnson2023mimic,hager_evaluation_2024}. All analyses were conducted on de-identified data, and no directly identifiable information was used for model evaluation, reporting or dissemination.

The external institution cohort from the Chinese PLA General Hospital comprised retrospectively collected encounters and was de-identified prior to analysis under institutional policies. Use of these records for this study was reviewed and approved by the hospital’s institutional ethics committee of the Chinese PLA General Hospital (Approval No. S2020-418-01), with a waiver of informed consent where applicable under the approved protocol. Data access was authorized through institutional governance procedures, and all processing and analyses were performed by authorized study personnel within institutionally approved computing environments.


\subsection{Models and implementation}\label{subsec:methods-models}

DxEvolve was implemented as an LLM-orchestrated agent operating in a workflow-aligned diagnostic environment with a constrained action schema, standardized tool interfaces and explicit termination criteria. Across all experiments, we used off-the-shelf, open-weight base LLMs. Model inference was conducted locally to satisfy data-governance requirements for both the MIMIC-derived benchmark and the external hospital cohort, which preclude transmitting patient-level content to third-party hosted LLM services or external APIs.

\textbf{Base LLMs and inference settings.}
Unless otherwise stated, all experiments in this study applied Qwen3-30B (\texttt{Qwen3-30B-A3B-Instruct}), Qwen3-235B (\texttt{Qwen3-235B-A22B-Instruct-2507})~\cite{yang2025qwen3}, \texttt{DeepSeek-V3.2}~\cite{liu2025deepseek} and \texttt{GLM-4.7}~\cite{5team2025glm45agenticreasoningcoding} as backbones. To contextualize DxEvolve against domain-specific models, we also evaluated MedGemma~\cite{sellergren2025medgemma} (\texttt{medgemma-27b-text-it}) and ClinicalCamel~\cite{toma2023clinical} (\texttt{ClinicalCamel-70B}). During preliminary testing, these medical-domain LLMs demonstrated insufficient compliance with the structured action-calling protocol required for workflow-aligned evaluation; specifically, they frequently failed to adhere to the pre-specified JSON output format or generated invalid investigative actions. Consequently, these models were evaluated exclusively under the single-pass full-information regime.
All experiments were run on a local server equipped with NVIDIA A100 GPUs (80\,GB), without using external hosted services. Within each base model, decoding configurations were held fixed across all compared methods and ablations to ensure that differences reflect workflow and experience mechanisms rather than sampling settings. For all evaluated LLMs, we set temperature to $0.1$, top-$p$ to $0.7$ and top-$k$ to $50$. 

\textbf{Prompt specification.}
All workflow-aligned experiments used a single, shared prompt contract that defines the action space and semantics, tool-call formatting, the agent state representation and the termination criteria. The same prompt template was applied across all evaluated base models without model-specific adapters or task-conditional modifications, ensuring that comparisons differ only in the underlying model and the enabled system components. Full prompt templates are provided in the Supplementary Section A and B.

\textbf{Retrieval settings.}
DxEvolve uses a unified dense retrieval stack for both (i) experience retrieval from the DCP repository and (ii) retrieval of external clinical guidelines when enabled. For both retrieval pathways, queries and candidate documents were embedded using \texttt{bge-large-en-v1.5}~\cite{bge_embedding} as dense encoder with vector-based similarity search (FAISS~\cite{johnson2019billion}). Similarity was computed by cosine similarity between $\ell_2$-normalized embeddings. Retrieval was performed locally for reproducibility and, for sensitive cohorts, to avoid external transfer of patient information. 
We collected abdominal-condition guideline documents from authoritative clinical sources (for example, the American College of Gastroenterology, the World Society of Emergency Surgery and Mayo Clinic) and manually verified relevance, authority and recency, excluding outdated materials and ultimately retained 35 guidelines. The guidelines were converted to structured text, lightly cleaned (for example, removing acknowledgements) before being locally indexed for retrieval.
PubMed retrieval was implemented via the official NCBI Entrez (E-utilities) API, with queries restricted to de-identified, non-patient-specific medical terms (for example, disease and symptom keywords) and containing no patient-level records or identifiable information.

\textbf{Baseline details and implementation parity.}
We use two complementary reference points: a published workflow-aligned baseline (CDM~\cite{hager_evaluation_2024}) and an in-framework ablation (DxEvolve w/o DCP) that isolates the marginal contribution of DCR and self-evolution mechanism. CDM is an established clinical decision-making diagnostic baseline capable of stepwise inquiry but lacking both a specialized investigative architecture for evidence acquisition and a framework for experiential evolution. 
Our evaluation strategy prioritizes head-to-head, backbone-matched ablations within a unified architectural framework, an approach designed to isolate the specific contributions of workflow grounding and experiential reuse. Direct comparisons with general-purpose agent frameworks are confounded by fundamental disparities in their underlying diagnostic paradigms. For instance, most existing models focus on exam-centric reasoning like USMLE-style scenarios, or are optimized for patient-physician dialogues. These settings diverge significantly from the sequential, uncertainty-laden investigation inherent to real-world clinical workups, where evidence is latent and must be actively requisitioned. To preserve domain fidelity, DxEvolve is intentionally architected to mirror the structured rigor of actual bedside practice, where evidence is latent and must be actively requisitioned. Such divergent information-access constraints and interaction modes make evaluation parity non-trivial; benchmarking against a standardized, workflow-aligned baseline and its corresponding ablations therefore ensures that observed gains are strictly attributable to our architectural innovations rather than artifacts of mismatched task definitions.

\subsection{Evaluation and analysis}
\label{subsec:methods-analysis}

This section defines the evaluation protocol and analysis definitions used throughout the study. We report encounter-level diagnosis accuracy under the DCR workflow, complemented by regime comparisons against single-pass full-information (FI) inference, exposure-indexed self-evolution analyses based on DCP accrual, and process-level metrics that characterize evidence-acquisition behaviour. All analyses were conducted on held-out evaluation cohorts with prespecified encounter-level definitions.

\textbf{Episodes, regimes and primary endpoint.}
Each diagnostic episode starts from the presenting complaint and limited initial context. The agent iteratively issues actions to request additional evidence and receives results only for requested items. Episodes terminate when the agent outputs a final primary diagnosis or reaches a prespecified maximum number of 20 interaction steps. The primary endpoint is encounter-level correctness of the final primary diagnosis; episodes that terminate without a valid diagnosis output are scored as incorrect. We report two regimes that differ only in information availability and interaction constraints. In the interactive regime, the agent must explicitly request evidence and may condition decisions only on evidence acquired within the episode. In single-pass full-information (FI) inference, the model receives the complete record upfront and produces a single-step diagnosis. Single-pass FI inference was evaluated only on the reader-study subset ($n$=80) as a matched control.

\textbf{Investigative burden and stratification.}
To analyze the efficacy of DxEvolve across varying levels of diagnostic difficulty, we defined an investigative complexity proxy derived from the baseline diagnostic burden. For each encounter, complexity was quantified as the evidence-acquisition footprint—defined as the total number of investigative steps required by the baseline CDM model to reach termination. Encounters were stratified into ``high-burden'' and ``low-burden'' groups based on a median split of this footprint across the 400-case evaluation cohort. This stratification allowed us to assess whether experience-guided evolution provides differential benefits in cases requiring extensive iterative reasoning versus more straightforward clinical presentations.


\textbf{Longitudinal self-evolution and improvement cases provenance.}
To quantify exposure-dependent self-evolution, we varied the number of encounters available for DCP accrual while holding the evaluation cohort fixed ($n$=400). Accrual encounters were ordered deterministically, and DCP repositories were constructed in a nested manner: at exposure level $k$, the repository contains DCPs consolidated from the first $k$ accrual encounters. This design yields an exposure-indexed learning curve without repeated re-sampling. The DCP-free ablation (DxEvolve w/o DCP) is exposure-independent by construction and was evaluated under the same interactive constraints as a reference.

To isolate evaluation encounters in which DCP reuse plausibly contributes to error correction, we defined improvement cases as evaluation encounters satisfying all of the following criteria: (i) DxEvolve produced a correct primary diagnosis, (ii) DxEvolve w/o DCP produced an incorrect diagnosis under the same workflow constraints, and (iii) DxEvolve retrieved at least one DCP during the episode. For provenance analyses, each retrieved DCP was labeled by the outcome of its source accrual episode at the time of consolidation (correct versus incorrect primary diagnosis). We quantified provenance enrichment by comparing the distribution of source-episode outcomes among DCPs retrieved in improvement cases against the corresponding distribution among DCPs retrieved across the full evaluation cohort (that is, pooling retrieval events over all evaluation encounters). Unless otherwise stated, provenance analyses were performed using the fixed accrual pool defined by the non-overlapping MIMIC-CDM split.

\textbf{Clinician assessment of DCP clinical maturation.}
To assess whether DCPs consolidated later in exposure are more clinically useful and reusable, we conducted a clinician reader study contrasting an early exposure window (encounters 1--300) and a late exposure window (encounters 1700--2000). For this assessment, we recruited two board-certified internal medicine physicians, one from the Chinese PLA General Hospital, China (with 15 years of clinical experience), one from the Peking University Third Hospital, China (with 8 years of clinical experience). Clinicians were masked to the exposure window of each DCP and the study hypothesis. From each window, we randomly sampled 20 DCPs (40 total). Each DCP was presented in its native three-part format (experience pattern, test-ordering experience and diagnostic decision experience) with all provenance metadata removed (including exposure index, source outcome and pathology labels) and translated to Chinese via a standardized translation procedure followed by terminology checks. Two board-certified clinicians independently rated each DCP on a 1--5 ordinal scale across three prespecified dimensions: clinical correctness (including potential safety concerns), actionability (capacity to guide evidence acquisition and hypothesis refinement in an interactive workflow) and generality (reusability beyond the originating encounter and pathology). Rating order was randomized and raters were blinded to sampling window and DCP source.
Inter-rater agreement for the clinician ratings was assessed using ordinal-appropriate reliability metrics (quadratic-weighted Cohen’s $\kappa$ and intraclass correlation). Agreement for the aggregate DCP score (mean across the three dimensions) was high (weighted $\kappa{=}0.83$, ICC$=0.81$), supporting the reliability of the clinician assessment for downstream analyses. For analysis and visualization, ratings were aggregated by averaging the two clinicians’ scores for each dimension and for the aggregate score.

\textbf{Process-level behaviour.}
We assessed evidence-acquisition behaviour by comparing the investigations requested by each method (DxEvolve and the CDM baseline) with those documented in the MIMIC-CDM structured record for the same encounters ($n$=400). All metrics were averaged across encounters.

$\bullet$ \emph{Trajectory consistency.} We quantified workup consistency using four complementary measures. (i) Physical examination (PE) agreement was a binary indicator of whether the agent requested a physical examination at any point in the episode (1 if requested, 0 otherwise). (ii) Laboratory-set F1 compared the set of laboratory tests ordered by the agent with the set recorded in MIMIC-CDM using a set-level F1 score. Before scoring, laboratory item identifiers were canonicalized using a precomputed mapping that collapses equivalent codes to a canonical identifier, reducing artefactual disagreement due to coding variations. Precision reflects avoidance of unnecessary tests, whereas recall reflects coverage of recorded tests. (iii) Imaging-set F1 was computed analogously, but over sets of (modality, region) tuples extracted from imaging requests, and a match required agreement on both modality and region. (iv) Action-order concordance evaluated whether the relative ordering of broad investigation types followed the reference clinical ordering. We restricted comparison to the intersection of investigation types executed by both the agent and the record; if fewer than two types were present, concordance was defined as 1. Otherwise, we computed pairwise concordance as the fraction of ordered pairs $(a,b)$ consistent with the reference order that were also ordered as $a$ before $b$ in the agent’s episode.

$\bullet$ \emph{Clinical guideline adherence proxies.} We additionally scored adherence to guideline-informed workup expectations using rules-based proxies with three components, reported on a 0–100 scale and averaged to form an overall score. (i) PE timing score captured whether PE was performed as the first workup step (100), performed later (50) or not performed (0). (ii) Laboratory adherence score measured coverage of pathology-specific recommended laboratory categories with a two-tier weighting scheme: primary tests contributed weight 1.0 each, secondary tests contributed weight 0.5 each with the total secondary contribution capped by the primary maximum to prevent inflation by extensive secondary testing; scores were normalized by the maximum attainable weight for the pathology. (iii) Imaging adherence score evaluated only the first imaging study, scoring whether its modality and region matched a pathology-specific preferred option (100), an acceptable alternative (50) or otherwise (0), including missing imaging. Guideline categories and imaging preferences were derived from established society guidelines (WSES~\cite{di2020diagnosis, sartelli20202020, leppaniemi20192019} for appendicitis, diverticulitis and pancreatitis; Tokyo Guidelines~\cite{yokoe2018tokyo} for cholecystitis), and this analysis was intended as a conservative, descriptive check for gross deviations rather than a claim of a single optimal workup for all clinical contexts.

%% file: secs/supplement.tex
\setcounter{page}{1}  
\renewcommand{\thepage}{S\arabic{page}} 

\begin{center}
\Large\bfseries Supplementary Information
\end{center}

\section*{A\quad Diagnostic Prompt Template}\label{secA1}

The following is the main diagnostic prompt template of DxEvolve used in all experiments across various base models reported in this paper, with medical examinations, experience retrieval, clinical guidelines, and PubMed search enabled.
Template variables are shown in \texttt{\{braces\}}.
The tags \texttt{\{system\_tag\_start\}}, \texttt{\{system\_tag\_end\}}, \texttt{\{user\_tag\_start\}}, \texttt{\{user\_tag\_end\}}, and \texttt{\{ai\_tag\_start\}} are replaced with model-specific chat delimiters at runtime.


\begin{longtable}{@{}p{\dimexpr\textwidth-2\tabcolsep}@{}}
\caption{Diagnostic Prompt Template.} \label{tab:diagnostic_prompt} \\
\toprule
\endfirsthead

\midrule
\endhead

\midrule
\multicolumn{1}{r}{\textit{Continued on next page}} \\
\endfoot

\bottomrule
\endlastfoot

\texttt{\{system\_tag\_start\}} \\
You are a senior physician. Your task is to perform stepwise diagnostic reasoning using ONLY the allowed tools. You must strictly follow one of the two output formats below at every step. \\
\\
FORMAT A. INFORMATION GATHERING \\
Thought: {[}1-2 concise sentences: what you know + what uncertainty remains + why next action is needed{]} \\
Action: {[}One of: Physical Examination, Laboratory Tests, Imaging, Experience Search, Guideline Search, PubMed Search{]} \\
Action Input: {[}Specific and valid request, MUST be within tool scope{]} \\
Observation: \\
{[}The system will fill this. DO NOT include any results yourself.{]} \\
\\
FORMAT B. FINAL DIAGNOSIS \\
Thought: {[}1-2 concise sentences summarizing key findings leading to the diagnosis{]} \\
Final Diagnosis: {[}Single, clear, concise, and standard diagnosis. (Avoid overly complex or speculative etiological chains, focus on the most likely and commonly recognized diagnosis.){]} \\
\\
STRICT RULES: \\
1. You MUST always follow the exact format (A or B). No deviations. \\
2. For any test, ONLY request those allowed by the corresponding tool. \\
\quad - Laboratory Tests: only valid lab names. \\
\quad - Imaging: must specify `\textless{}REGION\textgreater{} \textless{}MODALITY\textgreater{}' format (e.g., `Abdomen Ultrasound', `Abdomen CT'). \\
\quad - No invented tests, no unsupported modalities. \\
3. Before giving the final diagnosis, you MUST explicitly perform all three core types of medical evaluation as actions -- at least one Physical Examination, one Laboratory Test, and one Imaging. \\
\quad - Consider all clinically relevant imaging modalities for the suspected condition. \\
\quad - Do not omit a modality that is commonly recommended or diagnostically critical unless it is clearly inappropriate. \\
4. You MUST use Experience Search at least once before giving the final diagnosis. \\
\quad - In Action Input you SHOULD provide a short case style description of this patient (age, sex, chief complaint, symptom pattern, duration, key exam or lab or imaging findings), not just a single disease keyword. \\
\quad - If the retrieved experience is clearly irrelevant or not useful, you may reformulate the Action Input once and try a second Experience Search query. Do NOT keep searching repeatedly. \\
\quad - Only integrate insights that are consistent with this patient's objective data. \\
5. You MUST use Guideline Search at least once before giving the final diagnosis. \\
6. Stop when a confident diagnosis is possible based on available information. \\
7. When using Experience Search, Guideline Search, or PubMed Search, integrate only relevant insights into your Thought and proceed; do not rely on them if they conflict with patient-specific objective data. \\
8. If uncertainty remains but no high-yield action exists, you MUST provide the best-supported diagnosis (Format B) based on currently available data, without loop actions indefinitely. \\
\\
CRITICAL FORMAT RULES: \\
1. MUST output the "Observation:" label immediately after Action Input as a signal to pause for respond. \\
2. Keep "Action", "Action Input" and "Final Diagnosis" fields concise and to the point. \\
\\
AVAILABLE TOOLS: \\
- Physical Examination: Request physical examination of patient and receive the observations. This is a strongly recommended Examination in the clinical diagnostic process and should be performed first. \\
- Laboratory Tests: Request specific laboratory test and receive text values. Specify test names in 'Action Input' clearly. This is a common diagnostic step in the clinical evaluation. \\
- Imaging: Request imaging scans and receive the radiologist report. Region AND modality MUST be specified in the 'Action Input' field. \\
- Experience Search: Dense retrieval over past diagnostic cases. Action Input SHOULD be a short case style description of this patient, not just a disease name. \\
- Guideline Search: Retrieve relevant clinical guidelines. Provide a concise clinical query in "Action Input" (symptoms, suspected diagnosis, key labs/imaging, or decision point). \\
- PubMed Search: Conduct targeted search on PubMed and receive relevant medical articles. Concise and specific search query (few KEYWORDS) MUST be specified in "Action Input". \\
\\
BE EFFICIENT: Prioritize high-yield diagnostic actions before broad or low-yield ones. Some medical examination information may not be available, do not focus on the unavailable data, make full use of the information that can be obtained to diagnose. \\
\texttt{\{system\_tag\_end\}\{user\_tag\_start\}} \\
\\
Patient History: \\
\texttt{\{input\}} \\
\\
BEGIN YOUR DIAGNOSTIC PROCESS: \\
\texttt{\{user\_tag\_end\}\{ai\_tag\_start\}} \\
Thought:\texttt{\{agent\_scratchpad\}} \\
\end{longtable}

\noindent\textbf{Annotation.}
The prompt instructs the LLM to act as a senior physician performing stepwise diagnostic reasoning in an action-based loop.
Two output formats are enforced: Format~A for iterative information gathering (Thought $\to$ Action $\to$ Observation) and Format~B for the final diagnosis with thought.

\section*{B\quad Experience Construction Prompt Template}\label{secA2}

After each diagnostic case is completed, the following template is used to distill the case into a reusable diagnostic cognition primitive (DCP) through reflection on the diagnostic trajectory.
The DCP is stored in the DCP repository for retrieval in future cases.


\begin{longtable}{@{}p{\dimexpr\textwidth-2\tabcolsep}@{}}
\caption{Experience Construction Prompt.} \label{tab:experience_construction} \\
\toprule
\endfirsthead

\midrule
\endhead

\midrule
\multicolumn{1}{r}{\textit{Continued on next page}} \\
\endfoot

\bottomrule
\endlastfoot

\texttt{\{system\_tag\_start\}} \\
You extract reusable diagnostic reasoning experience from completed clinical cases for future tool using agents. \\
\\
Your goal: \\
- Do NOT retell the full case or reproduce chain of thought. \\
- Do NOT include treatment. \\
- Distill ONE Diagnostic Cognition Primitive (DCP): a short heuristic that improves future diagnosis. \\
\\
The DCP must: \\
- Be consistent with the ground\_truth diagnosis and the correctness flag. \\
- Focus on diagnostic reasoning, not management or consultation. \\
- Emphasize when and how to use ONLY the following tools in future similar cases: \\
\quad - Physical Examination (no additional input) \\
\quad - Laboratory Tests (input: names of the lab tests to run) \\
\quad - Imaging (input: imaging modality and region to be scanned) \\
\\
Tool input templates (copyable): \\
- Physical Examination \\
- Laboratory Tests: \textless{}test name 1\textgreater{}, \textless{}test name 2\textgreater{}, ... \\
- Imaging: modality=\textless{}MODALITY\textgreater{}, region=\textless{}REGION\textgreater{} \\
\\
Coverage constraints: \\
- Only recommend tests or imaging settings that are explicitly supported by the provided case context, meaning they appear in at least one of: \\
\quad 1) Clinician test orders (from the chart). Use this as a high quality reference for realistic first line test selection and sequencing. \\
\quad 2) Diagnostic steps where the tool call succeeded (has a non-error observation) \\
\quad 3) Rule based feedback `message' or retrieved guidance that explicitly recommends a specific test or imaging setting \\
- Prefer to fully cover the explicitly provided clinician orders and successful tool calls before adding anything else. \\
- Do not invent new tests, imaging modalities, regions, or non-provided measurement names. \\
\\
Field roles: \\
- Experience Pattern: \\
\quad - Case-style trigger pattern for retrieval, built from symptoms, basic context, and key objective findings. \\
\quad - You may append compact labels such as the final correct diagnosis and common misdiagnoses to improve retrieval. \\
- Test Ordering Experience: \\
\quad - Constructive test-ordering heuristic using only the allowed tools and tool-compatible inputs. \\
\quad - You may rank actions by priority and specify escalation criteria, in natural clinical language. \\
\quad - Avoid blanket prohibitions. If a test is lower priority, express it as conditional or deferred rather than discouraged. \\
\quad - When naming tests or imaging, use the copyable tool input templates above. \\
- Diagnostic Decision Experience: \\
\quad - Short rule on how to weigh key findings and move from differential diagnosis to the correct final diagnosis. \\
\\
Error correction rules: \\
- If correctness is "Correct": \\
\quad - Treat the model's diagnostic process as broadly appropriate. \\
\quad - Extract the most reusable diagnostic pattern and test ordering heuristic. \\
- If correctness is "Incorrect": \\
\quad - Treat the model's final diagnosis and reasoning as a negative example. \\
\quad - Do NOT justify or reuse the incorrect diagnosis. \\
\quad - Use the ground\_truth and the rule based feedback in `message' as the primary reference. \\
\quad - Base the DCP on the ideal diagnostic process implied by that feedback. \\
\\
Input fields: \\
- Patient input: raw case description. \\
- Diagnostic steps: chronological list of tool calls and observations. \\
- Model final diagnosis: what the model concluded. \\
- Ground truth diagnosis: correct diagnosis label for this case. \\
- Correctness flag: "Correct" or "Incorrect". \\
- Rule based feedback: comments about missing exams, unnecessary tests, wrong imaging, and efficiency. \\
- Clinician test orders (from the chart): tests ordered by the treating clinician as documented in the chart, expressed with the same tool names and inputs, and serving as a realistic reference for first line test selection and sequencing. \\
\\
Case context: \\
Patient input: \\
\texttt{\{input\}} \\
\\
Diagnostic steps: \\
\texttt{\{intermediate\_steps\}} \\
\\
Model final diagnosis: \\
\texttt{\{output\}} \\
\\
Ground truth diagnosis: \\
\texttt{\{ground\_truth\}} \\
\\
Correctness flag: \\
\texttt{\{correctness\}} \\
\\
Rule based feedback on process: \\
\texttt{\{message\}} \\
\\
Clinician test orders (from the chart): \\
\texttt{\{clinician\}} \\
\\
Now output exactly in this format: \\
\\
Experience Pattern: \textless{}case-style trigger pattern for retrieval. may include final correct diagnosis and frequent misdiagnoses as compact labels\textgreater{} \\
Test Ordering Experience: \textless{}priority ranked ordering sequence and escalation criteria, using the tool input templates and respecting the coverage constraints\textgreater{} \\
Diagnostic Decision Experience: \textless{}concise rule on how to weigh key findings and reach the correct diagnosis, aligned with ground\_truth and rule based feedback\textgreater{} \\
\texttt{\{system\_tag\_end\}} \\
\\
\texttt{\{ai\_tag\_start\}} \\
\end{longtable}

\noindent\textbf{Annotation.}
This template implements the \emph{Experience Construction} module that generates DCPs from completed cases.
Each DCP consists of three fields: (1)~{Experience Pattern}, a case-style trigger description optimized for dense retrieval; (2)~{Test Ordering Experience}, a prioritized test-ordering heuristic grounded in clinician orders and successful tool calls; and (3)~{Diagnostic Decision Experience}, a concise rule for weighing findings toward the correct diagnosis.
The \texttt{\{message\}} variable contains rule-based evaluator feedback on the diagnostic process, which identifies missing examinations, unnecessary tests, or procedural deviations based on pathology-specific evaluation criteria. For example, if the agent failed to request appropriate imaging for suspected appendicitis, the feedback might state: ``Imaging: no appropriate abdominal imaging was requested. Set region=`Abdomen' and request imaging (ultrasound is typically preferred in pediatric or pregnant patients, while CT is generally recommended for adult non-pregnant patients).'' This feedback guides the DCP construction to emphasize the correct diagnostic workflow.
The \texttt{\{clinician\}} variable provides real clinician test orders extracted from the MIMIC-IV chart, serving as a high-quality reference for realistic test selection and sequencing.

\section*{C\quad Example Diagnostic Cognition Primitive}\label{secA3}

The following is a representative DCP generated through reflection on the diagnostic trajectory from a correctly diagnosed case of acute biliary pancreatitis.
This DCP is stored in the DCP repository and retrieved via vector-based dense retrieval when the agent encounters similar presentations in future cases.


\begin{longtable}{@{}p{\dimexpr\textwidth-2\tabcolsep}@{}}
\caption{Example DCP (Correct Case).} \label{tab:dcp_correct} \\
\toprule
\endfirsthead

\midrule
\endhead

\midrule
\multicolumn{1}{r}{\textit{Continued on next page}} \\
\bottomrule
\endfoot

\bottomrule
\endlastfoot

Experience Pattern: \\
Post-cholecystectomy patient with acute RUQ/back pain, elevated liver enzymes and lipase. (Acute pancreatitis, DDx: Biliary pancreatitis vs. other etiologies) \\
\\
Test Ordering Experience: \\
First, confirm pancreatitis with Laboratory Tests: Lipase, Amylase, CBC, CMP. Concurrently, order first-line biliary imaging: Imaging: modality=Ultrasound, region=Abdomen. \\
If ultrasound is negative for stones/dilation but liver enzymes remain elevated, escalate to definitive biliary evaluation (ERCP) per clinician orders; do not escalate to CT or MRCP without specific indications (e.g., concern for complications or failed ERCP). \\
\\
Diagnostic Decision Experience: \\
In a post-cholecystectomy setting, acute pancreatitis with concurrent transaminitis/hyperbilirubinemia is biliary in origin until proven otherwise, even with a negative initial ultrasound, as microlithiasis or sphincter dysfunction may be the cause. \\
\end{longtable}

\medskip
Below is a second example DCP generated from an incorrectly diagnosed case, demonstrating the error correction mechanism.
The agent originally diagnosed ``adhesive small bowel obstruction'' but the ground truth was cholecystitis.


\begin{longtable}{@{}p{\dimexpr\textwidth-2\tabcolsep}@{}}
\caption{Example DCP (Incorrect Case).} \label{tab:dcp_incorrect} \\
\toprule
\endfirsthead

\midrule
\endhead

\midrule
\multicolumn{1}{r}{\textit{Continued on next page}} \\
\bottomrule
\endfoot

\bottomrule
\endlastfoot

Experience Pattern: \\
Young to middle-aged female with acute right abdominal pain, sharp on palpation, bilious vomiting, chills, and history of prior abdominal surgeries (e.g., laparoscopies). Past medical history of endometriosis. (Correct: cholecystitis; Common misdiagnosis: adhesive small bowel obstruction) \\
\\
Test Ordering Experience: \\
1. Physical Examination. \\
2. Laboratory Tests: CBC differential, CMP, (Blood) Lactate, (Urine) HCG. \\
3. Imaging: modality=Ultrasound, region=Abdomen. \\
Escalate to further imaging (e.g., CT) only if ultrasound is non-diagnostic and clinical suspicion for obstruction or other complication remains high. \\
\\
Diagnostic Decision Experience: \\
In a patient with right upper quadrant or right-sided abdominal pain, vomiting, and chills, prioritize gallbladder pathology. A history of prior surgery should not prematurely anchor to adhesive obstruction; a finding of gallstones on ultrasound, especially with local tenderness, strongly supports cholecystitis over obstruction. \\
\end{longtable}

\noindent\textbf{Annotation.}
The first DCP illustrates how a correctly diagnosed case is consolidated into a reusable experience artifact. The {experience pattern} provides a high-salience signature for retrieval, summarizing the presentation and discriminative cues. The {test-ordering experience} encodes actionable workup guidance, including high-yield next-step evaluations and contingency options. The {diagnostic decision experience} captures evidence-linked implications for hypothesis refinement and final decision-making.
The second DCP demonstrates how corrective lessons are incorporated when the source trajectory exposed an error mode: when the agent misdiagnosed cholecystitis as small bowel obstruction in a case with atypical presentation, the DCP was constructed from the ground truth and evaluator feedback, explicitly labeling the common misdiagnosis and providing the correct reasoning pathway.

\section*{D\quad Example Diagnostic Reasoning Trace}\label{secA4}

The following is a complete diagnostic reasoning trace from a real case in the MIMIC-CDM benchmark, showing the agent's stepwise process from initial presentation to final diagnosis.
The case involves an elderly patient with diabetes presenting with acute right upper quadrant abdominal pain, ultimately diagnosed with acute calculous cholecystitis.


\begin{longtable}{@{}p{\dimexpr\textwidth-2\tabcolsep}@{}}
\caption{Example Diagnostic Reasoning Trace.} \label{tab:diagnostic_trace} \\
\toprule
\endfirsthead

\midrule
\endhead

\midrule
\multicolumn{1}{r}{\textit{Continued on next page}} \\
\endfoot

\bottomrule
\endlastfoot

Patient History: \\
\\
Elderly male patient with history of type 2 diabetes mellitus on insulin, hypothyroidism, hypertension, and prostate cancer status-post radiotherapy presented to the emergency department with acute onset abdominal pain. The patient reported that the pain began suddenly at approximately 3 AM, waking him from sleep. He described it as sharp, constant, and localized to the right side of the abdomen. When the pain persisted, he initially attempted to contact his primary care physician but was unable to secure an appointment, prompting him to call emergency services. \\
\\
In the emergency department, initial vital signs were temperature 98°F, blood pressure 150/86 mmHg, heart rate normal, respiratory rate 16 breaths per minute, oxygen saturation 100\% on room air. Initial laboratory studies showed lactate 2.3 mmol/L, glucose 279 mg/dL, normal liver function tests, and hematocrit 33.2\%. Right upper quadrant ultrasound demonstrated extrahepatic biliary duct dilation with common bile duct measuring 10 mm and presence of gallstones. Computed tomography of the abdomen and pelvis confirmed these findings. He was treated with intravenous fluids, morphine for pain control, and ondansetron for nausea. \\
\\
On admission to the hospital floor, the patient reported improved but persistent right upper quadrant abdominal pain. He endorsed dry heaves and constipation, with his last bowel movement occurring two days prior. He denied fever, chills, chest pain, or shortness of breath. \\
\\
Past Medical History: Prostate cancer status-post radiotherapy, type 2 diabetes mellitus on insulin, hypertension, hypothyroidism, chronic kidney disease stage III, history of gallstones, nephrolithiasis, osteoarthritis, glaucoma, cataracts, sickle cell trait, hearing loss. \\
\\
Past Surgical History: Pilonidal cyst excision, bilateral inguinal hernia repairs. \\
\\
Social History: Details not specified. \\
\\
Family History: Type 2 diabetes mellitus in family members; no family history of prostate cancer. \\
\\
\\
Step 1: Physical Examination \\
\\
Action: Physical Examination \\
\\
Observation: \\
Vitals: Temperature 98.2°F, BP 148/68 mmHg, HR 90 bpm, RR 20, O2 sat 100\%. \\
\\
General: Alert and oriented, no acute distress. \\
\\
HEENT: Sclerae anicteric, moist mucous membranes, oropharynx clear, pupils equal and reactive to light, bilateral cloudy lenses (cataracts). \\
\\
Cardiovascular: Regular rate and rhythm, normal S1 and S2 heart sounds, no murmurs, rubs, or gallops. \\
\\
Pulmonary: Clear to auscultation bilaterally, no wheezes, rales, or rhonchi. \\
\\
Abdomen: Soft, tender to palpation in the right upper quadrant, non-distended, bowel sounds present, no organomegaly, no rebound tenderness or guarding. \\
\\
Genitourinary: No Foley catheter in place. \\
\\
Extremities: Warm and well-perfused, 2+ pulses bilaterally, no clubbing, cyanosis, or edema. \\
\\
Neurological: Cranial nerves II-XII grossly intact, strength grossly intact, noted hearing loss. \\
\\
\\
Step 2: Laboratory Tests \\
\\
Action: Laboratory Tests \\
\\
Action Input: Complete blood count with differential, comprehensive metabolic panel, lipase \\
\\
Observation: \\
White Blood Cell Count: 7.6 K/uL (reference range: 4.0--11.0 K/uL) \\
\\
Neutrophils: 89.1\%, Absolute Neutrophil Count: 6.77 K/uL {[}Elevated, indicating neutrophilia{]} \\
\\
Lymphocytes: 6.4\%, Absolute Lymphocyte Count: 0.49 K/uL \\
\\
Monocytes: 3.2\%, Absolute Monocyte Count: 0.24 K/uL \\
\\
Eosinophils: 0.4\%, Absolute Eosinophil Count: 0.03 K/uL \\
\\
Basophils: 0.5\%, Absolute Basophil Count: 0.04 K/uL \\
\\
Red Blood Cell Count: 4.73 M/uL \\
\\
Hemoglobin: 11.2 g/dL (reference range: 13.5--17.5 g/dL) {[}Low{]} \\
\\
Hematocrit: 33.2\% (reference range: 41--53\%) {[}Low{]} \\
\\
MCV: 70.0 fL (reference range: 80--100 fL) {[}Low, microcytic{]} \\
\\
MCH: 23.7 pg (reference range: 27--31 pg) {[}Low{]} \\
\\
MCHC: 33.7 g/dL (reference range: 32--36 g/dL) \\
\\
Platelet Count: 156 K/uL (reference range: 150--400 K/uL) \\
\\
RDW: 14.6\% (reference range: 11.5--14.5\%) \\
\\
Glucose: 279 mg/dL (reference range: 70--100 mg/dL) {[}Elevated{]} \\
\\
Sodium: 141 mEq/L (reference range: 136--145 mEq/L) \\
\\
Potassium: 3.6 mEq/L (reference range: 3.5--5.1 mEq/L) \\
\\
Chloride: 99 mEq/L (reference range: 98--107 mEq/L) \\
\\
Bicarbonate: 25 mEq/L (reference range: 22--29 mEq/L) \\
\\
Blood Urea Nitrogen: 19 mg/dL (reference range: 7--20 mg/dL) \\
\\
Creatinine: 1.1 mg/dL (reference range: 0.7--1.3 mg/dL) \\
\\
Calcium, Total: 9.3 mg/dL (reference range: 8.5--10.5 mg/dL) \\
\\
Alanine Aminotransferase (ALT): 43 IU/L (reference range: 7--56 IU/L) \\
\\
Aspartate Aminotransferase (AST): 30 IU/L (reference range: 10--40 IU/L) \\
\\
Alkaline Phosphatase: 91 IU/L (reference range: 40--150 IU/L) \\
\\
Total Bilirubin: 0.6 mg/dL (reference range: 0.3--1.2 mg/dL) \\
\\
Lipase: 12 IU/L (reference range: 7--60 IU/L) \\
\\
\\
Step 3: Imaging (Abdominal Ultrasound) \\
\\
Action: Imaging \\
\\
Action Input: Abdomen Ultrasound (focused on gallbladder and biliary tree) \\
\\
Observation: \\
Right Upper Quadrant Ultrasound: \\
\\
TECHNIQUE: Grayscale and color Doppler ultrasound images of the abdomen. \\
\\
FINDINGS: \\
\\
Liver: Homogenous attenuation throughout. No focal lesions identified. \\
\\
Hepatobiliary: Mild intrahepatic and extrahepatic biliary duct dilation noted. Common bile duct measures 12 mm in diameter. Gallbladder contains gallstones without wall thickening or surrounding inflammation visible on this initial study. Trace pericholecystic free fluid noted. \\
\\
Pancreas: Diffuse pancreatic atrophy. No pancreatic duct dilation or peripancreatic fat stranding. \\
\\
Spleen: Normal size and attenuation, no focal lesions. \\
\\
Kidneys: Bilateral subcentimeter renal hypodensities too small to characterize. A 6.1 × 5.6 cm simple cyst in the interpolar region of the left kidney. No hydronephrosis or perinephric abnormality. \\
\\
IMPRESSION: Cholelithiasis, common bile duct dilation (12 mm), trace pericholecystic fluid. Recommend correlation with clinical findings and consider advanced biliary imaging if indicated. \\
\\
Step 4: Experience Search \\
\\
Action: Experience Search \\
\\
Action Input: Elderly male with diabetes, acute RUQ pain, leukocytosis, normal LFTs, imaging shows acute cholecystitis (gallbladder wall edema, pericholecystic fluid) and common bile duct dilation with papillary stenosis from passed stones \\
\\
Observation: \\
Case 1: \\
\\
Experience Pattern: Elderly female with known gallstones and prior biliary intervention (ERCP/stent) presents with acute right upper quadrant pain and tenderness; normal white blood cell count and liver enzymes; ultrasound shows gallstones, gallbladder wall thickening, and common bile duct dilation. {[}Correct: Acute Cholecystitis | Misdiagnoses: Cholangitis, Biliary Colic, Pancreatitis{]} \\
\\
Test Ordering Experience: First, perform a Physical Examination focusing on abdominal signs. Then, order initial Laboratory Tests: CBC, CMP (including liver enzymes and bilirubin), and lipase. If these are normal but suspicion remains, escalate to Laboratory Tests: C-reactive protein (CRP). For imaging, the first and often definitive test is Imaging: modality=Ultrasound, region=Abdomen (specifically right upper quadrant). Do not escalate to CT or MRCP unless the ultrasound is non-diagnostic or complications are suspected. \\
\\
Diagnostic Decision Experience: In a patient with known gallstones, acute RUQ pain/tenderness, and ultrasound findings of gallstones with gallbladder wall thickening, diagnose acute cholecystitis even if inflammatory markers and LFTs are normal; CBD dilation and pneumobilia in this context are likely chronic sequelae of prior ERCP and do not rule out cholecystitis. \\
\\
{[}Additional guidelines omitted for brevity{]} \\
\\
Step 5: Imaging (MRCP) \\
\\
Action: Imaging \\
\\
Action Input: Abdomen MRCP \\
\\
Observation: \\
Magnetic Resonance Cholangiopancreatography (MRCP): \\
\\
TECHNIQUE: T1- and T2-weighted multiplanar images of the abdomen acquired in a 1.5 T magnet. Intravenous contrast: 9 mL Gadavist. Oral contrast: 1 cc of Gadavist mixed with 50 cc of water. \\
\\
FINDINGS: \\
\\
Lower Thorax: Mild bibasilar atelectasis. No large pleural effusion. \\
\\
Liver: Homogeneous signal intensity without focal lesions. Mild, nonspecific periportal edema. \\
\\
Biliary: The gallbladder is distended and contains numerous calculi. Since prior examinations, there is progressive gallbladder wall edema and pericholecystic fluid with gallbladder wall enhancement. No clear pericholecystic hepatic hyperenhancement. The common bile duct measures up to 1.2 cm in diameter. It tapers abruptly with a fixed loss in caliber, most consistent with papillary stenosis, likely the result of previously passed stones. \\
\\
Pancreas: Normal signal and enhancement. Several subcentimeter cystic lesions. \\
\\
Spleen: Normal size. \\
\\
Adrenal Glands: Unremarkable. \\
\\
Kidneys: Symmetric. Large simple cyst in the upper pole of the left kidney, unchanged. Minimal nonspecific perinephric fat stranding. \\
\\
Gastrointestinal Tract: No significant abnormalities. \\
\\
Lymph Nodes: No enlarged lymph nodes. \\
\\
Vasculature: Patent aorta and main branches. \\
\\
Osseous and Soft Tissue Structures: Mild degenerative changes throughout the lower thoracic and lumbar spine. Mild symmetric bilateral gynecomastia noted incidentally. \\
\\
IMPRESSION: Acute calculous cholecystitis with gallbladder wall edema, pericholecystic fluid, and wall enhancement. Common bile duct dilation with abrupt tapering suggestive of papillary stenosis from previously passed stones. \\
\\
Step 6: Guideline Search \\
\\
Action: Guideline Search \\
\\
Action Input: Management of acute cholecystitis with common bile duct dilation, suspected choledocholithiasis or papillary stenosis \\
\\
Observation: \\
Guideline 1 (score=0.5146) \\
\\
Title: 2020 World Society of Emergency Surgery updated guidelines for the diagnosis and treatment of acute calculus cholecystitis \\
\\
Year: 2020 \\
\\
Key points: \\
- Common bile duct dilation on ultrasound ($>$6 mm with gallbladder in situ) is a strong risk factor for common bile duct stones (CBDS), but is not diagnostic alone; further tests are required. \\
- For moderate-risk patients (e.g., CBD dilation + bilirubin 1.8--4 mg/dL), preoperative MRCP or endoscopic ultrasound is recommended to diagnose CBDS. \\
- For high-risk patients (e.g., visualized CBD stone on ultrasound or ascending cholangitis), preoperative ERCP is recommended to diagnose CBDS. \\
- An increased common bile duct diameter on ultrasound is an indirect sign but is not sufficient alone to diagnose choledocholithiasis, necessitating further diagnostic tests. \\
\\
{[}Additional guidelines omitted for brevity{]} \\
\\
Step 7: PubMed Search \\
\\
Action: PubMed Search \\
\\
Action Input: acute cholecystitis common bile duct dilation normal liver enzymes \\
\\
Observation: \\
Article 1: Role of liver function tests in symptomatic cholelithiasis. \\
\\
Abstract: Cholelithiasis is a common problem in Western and developing nations with continuously rising incidence. This prospective study evaluated the usefulness of routine liver function tests (LFTs) in patients undergoing cholecystectomy for symptomatic cholelithiasis. Out of 124 patients, 87.10\% had LFTs within normal range. Most patients with normal LFTs (95.37\%) presented with chronic calculous cholecystitis. Of those with deranged LFTs, 43.75\% presented with acute calculous cholecystitis. None had dilated common bile duct on preoperative ultrasound or intraoperative findings. Conclusion: Routine LFTs in preoperative assessment of uncomplicated symptomatic cholelithiasis usually come out normal and are not good predictive markers for detecting silent common bile duct calculi. \\
\\
Article 2: Endoscopic Retrograde Cholangiopancreatography in Kidney Transplant Patients: Results From a Single Center. \\
\\
Abstract: Twenty-one kidney transplant patients underwent 23 ERCP procedures for various biliary indications including choledocholithiasis (28.6\%), common bile duct dilation with liver enzyme elevations (19\%), and cholangitis (23.8\%). Findings included choledocholithiasis in 47.6\% and dilated common bile duct in 9.6\%. ERCP was safe with no complications such as acute pancreatitis, bleeding, or perforation. \\
\\
{[}Additional guidelines omitted for brevity{]} \\
\\
Final Diagnosis \\
\\
The patient presents with acute right upper quadrant pain, neutrophilia (89.1\% neutrophils, absolute neutrophil count 6.77 K/uL), and cholelithiasis on imaging. Initial ultrasound demonstrated gallstones with common bile duct dilation (12 mm) and trace pericholecystic fluid. MRCP definitively showed progressive gallbladder wall edema, pericholecystic fluid, and gallbladder wall enhancement—classic imaging findings for acute cholecystitis. The common bile duct dilation with abrupt tapering is consistent with papillary stenosis from previously passed stones. Notably, liver function tests remained entirely normal (ALT 43 IU/L, AST 30 IU/L, alkaline phosphatase 91 IU/L, bilirubin 0.6 mg/dL), and lipase was normal (12 IU/L), effectively ruling out concurrent pancreatitis or acute cholangitis. The retrieved clinical experiences and guidelines confirm that normal liver enzymes do not exclude acute cholecystitis, and that CBD dilation in this setting likely reflects chronic sequelae from stone passage rather than active choledocholithiasis. \\
\\
Final Diagnosis: Acute calculous cholecystitis \\
\\
Ground Truth: Cholecystitis \checkmark \\
\end{longtable}

\noindent\textbf{Annotation.}
This trace demonstrates the complete deep clinical research (DCR) workflow with actual laboratory results and rich experience retrieval.
The agent follows a clinically logical sequence: physical examination first revealing right upper quadrant tenderness, followed by comprehensive laboratory evaluation showing relative neutrophilia (89.1\% neutrophils, absolute neutrophil count 6.77 K/uL) with liver enzymes within normal limits (ALT 43 IU/L, AST 30 IU/L, alkaline phosphatase 91 IU/L, bilirubin 0.6 mg/dL) and normal lipase (12 IU/L).
Initial right upper quadrant ultrasound showed cholelithiasis with common bile duct dilation (12 mm) and trace pericholecystic fluid. The agent escalated to MRCP for more definitive biliary assessment, which revealed gallbladder wall thickening and edema, pericholecystic fluid, and increased T2 signal—findings consistent with acute calculous cholecystitis. 
The Experience Search retrieved relevant cases from the experience library, providing guidance on test-ordering strategies and diagnostic reasoning for similar presentations. The retrieved experiences noted that acute cholecystitis can present with normal liver enzymes and that CBD dilation in the absence of visualized stones reduces the likelihood of active choledocholithiasis.
The Guideline Search retrieved the 2020 World Society of Emergency Surgery guidelines on acute calculous cholecystitis, which informed the diagnostic reasoning regarding CBD dilation and the appropriateness of MRCP for moderate-risk patients.
The PubMed Search provided supporting evidence regarding the prevalence of normal liver function tests in acute cholecystitis.
The final diagnosis of acute calculous cholecystitis was correct, matching the ground truth label.

%% file: sn-bibliography.bib
@article{hager_evaluation_2024,
	title = {Evaluation and mitigation of the limitations of large language models in clinical decision-making},
	issn = {1546-170X},
	url = {https://doi.org/10.1038/s41591-024-03097-1},
	doi = {10.1038/s41591-024-03097-1},
    journal = {Nature Medicine},
	author = {Hager, Paul and Jungmann, Friederike and Holland, Robbie and Bhagat, Kunal and Hubrecht, Inga and Knauer, Manuel and Vielhauer, Jakob and Makowski, Marcus and Braren, Rickmer and Kaissis, Georgios and Rueckert, Daniel},
    year={2023},
}

@article{johnson2023mimic,
  title={MIMIC-IV, a freely accessible electronic health record dataset},
  author={Johnson, Alistair EW and Bulgarelli, Lucas and Shen, Lu and Gayles, Alvin and Shammout, Ayad and Horng, Steven and Pollard, Tom J and Hao, Sicheng and Moody, Benjamin and Gow, Brian and others},
  journal={Scientific data},
  volume={10},
  number={1},
  pages={1},
  year={2023},
  publisher={Nature Publishing Group UK London}
}

@article{li2025macd,
  title={MACD: Multi-Agent Clinical Diagnosis with Self-Learned Knowledge for LLM},
  author={Li, Wenliang and Yan, Rui and Zhang, Xu and Chen, Li and Zhu, Hongji and Zhao, Jing and Li, Junjun and Li, Mengru and Cao, Wei and Jiang, Zihang and others},
  journal={arXiv preprint arXiv:2509.20067},
  year={2025}
}

@article{gong2025knowledge,
  title={Knowledge-Practice Performance Gap in Clinical Large Language Models: Systematic Review of 39 Benchmarks},
  author={Gong, Eun Jeong and Bang, Chang Seok and Lee, Jae Jun and Baik, Gwang Ho},
  journal={Journal of Medical Internet Research},
  volume={27},
  pages={e84120},
  year={2025},
  publisher={JMIR Publications Toronto, Canada}
}

@article{mccoy2025assessment,
  title={Assessment of large language models in clinical reasoning: a novel benchmarking study},
  author={McCoy, Liam G and Swamy, Rajiv and Sagar, Nidhish and Wang, Minjia and Bacchi, Stephen and Fong, Jie Ming Nigel and Tan, Nigel CK and Tan, Kevin and Buckley, Thomas A and Brodeur, Peter and others},
  journal={NEJM AI},
  volume={2},
  number={10},
  pages={AIdbp2500120},
  year={2025},
  publisher={Massachusetts Medical Society}
}

@article{qiu2025quantifying,
  title={Quantifying the reasoning abilities of LLMs on clinical cases},
  author={Qiu, Pengcheng and Wu, Chaoyi and Liu, Shuyu and Fan, Yanjie and Zhao, Weike and Chen, Zhuoxia and Gu, Hongfei and Peng, Chuanjin and Zhang, Ya and Wang, Yanfeng and others},
  journal={Nature Communications},
  volume={16},
  number={1},
  pages={9799},
  year={2025},
  publisher={Nature Publishing Group UK London}
}

@article{singhal2023large,
  title={Large language models encode clinical knowledge},
  author={Singhal, Karan and Azizi, Shekoofeh and Tu, Tao and Mahdavi, S Sara and Wei, Jason and Chung, Hyung Won and Scales, Nathan and Tanwani, Ajay and Cole-Lewis, Heather and Pfohl, Stephen and others},
  journal={Nature},
  volume={620},
  number={7972},
  pages={172--180},
  year={2023},
  publisher={Nature Publishing Group}
}

@article{graber2005diagnostic,
  title={Diagnostic error in internal medicine},
  author={Graber, Mark L and Franklin, Nancy and Gordon, Ruthanna},
  journal={Archives of internal medicine},
  volume={165},
  number={13},
  pages={1493--1499},
  year={2005},
  publisher={American Medical Association}
}

@article{singh2014frequency,
  title={The frequency of diagnostic errors in outpatient care: estimations from three large observational studies involving US adult populations},
  author={Singh, Hardeep and Meyer, Ashley ND and Thomas, Eric J},
  journal={BMJ quality \& safety},
  volume={23},
  number={9},
  pages={727--731},
  year={2014},
  publisher={BMJ Publishing Group Ltd}
}

@book{ball2016improving,
  title={Improving Diagnosis in Health Care},
  author={Ball, John R and Miller, Bryan T and Balogh, Erin P},
  year={2016},
  publisher={National Academies Press}
}

@article{norman2017causes,
  title={The causes of errors in clinical reasoning: cognitive biases, knowledge deficits, and dual process thinking},
  author={Norman, Geoffrey R and Monteiro, Sandra D and Sherbino, Jonathan and Ilgen, Jonathan S and Schmidt, Henk G and Mamede, Silvia},
  journal={Academic Medicine},
  volume={92},
  number={1},
  pages={23--30},
  year={2017},
  publisher={LWW}
}

@misc{eriksen2024use,
  title={Use of GPT-4 to diagnose complex clinical cases},
  author={Eriksen, Alexander V and M{\"o}ller, S{\"o}ren and Ryg, Jesper},
  journal={Nejm Ai},
  volume={1},
  number={1},
  pages={AIp2300031},
  year={2024},
  publisher={Massachusetts Medical Society}
}

@article{achiam2023gpt,
  title={Gpt-4 technical report},
  author={Achiam, Josh and Adler, Steven and Agarwal, Sandhini and Ahmad, Lama and Akkaya, Ilge and Aleman, Florencia Leoni and Almeida, Diogo and Altenschmidt, Janko and Altman, Sam and Anadkat, Shyamal and others},
  journal={arXiv preprint arXiv:2303.08774},
  year={2023}
}

@article{savage2024diagnostic,
  title={Diagnostic reasoning prompts reveal the potential for large language model interpretability in medicine},
  author={Savage, Thomas and Nayak, Ashwin and Gallo, Robert and Rangan, Ekanath and Chen, Jonathan H},
  journal={NPJ Digital Medicine},
  volume={7},
  number={1},
  pages={20},
  year={2024},
  publisher={Nature Publishing Group UK London}
}

@article{han2024comparative,
  title={Comparative analysis of multimodal large language model performance on clinical vignette questions},
  author={Han, Tianyu and Adams, Lisa C and Bressem, Keno K and Busch, Felix and Nebelung, Sven and Truhn, Daniel},
  journal={JAMA},
  volume={331},
  number={15},
  pages={1320--1321},
  year={2024},
  publisher={American Medical Association}
}

@article{kaczmarczyk2024evaluating,
  title={Evaluating multimodal AI in medical diagnostics},
  author={Kaczmarczyk, Robert and Wilhelm, Theresa Isabelle and Martin, Ron and Roos, Jonas},
  journal={npj Digital Medicine},
  volume={7},
  number={1},
  pages={205},
  year={2024},
  publisher={Nature Publishing Group UK London}
}

@article{mcduff2025towards,
  title={Towards accurate differential diagnosis with large language models},
  author={McDuff, Daniel and Schaekermann, Mike and Tu, Tao and Palepu, Anil and Wang, Amy and Garrison, Jake and Singhal, Karan and Sharma, Yash and Azizi, Shekoofeh and Kulkarni, Kavita and others},
  journal={Nature},
  pages={1--7},
  year={2025},
  publisher={Nature Publishing Group UK London}
}

@article{zoller2025human,
  title={Human--AI collectives most accurately diagnose clinical vignettes},
  author={Z{\"o}ller, Nikolas and Berger, Julian and Lin, Irving and Fu, Nathan and Komarneni, Jayanth and Barabucci, Gioele and Laskowski, Kyle and Shia, Victor and Harack, Benjamin and Chu, Eugene A and others},
  journal={Proceedings of the National Academy of Sciences},
  volume={122},
  number={24},
  pages={e2426153122},
  year={2025},
  publisher={National Academy of Sciences}
}

@article{bhasuran2025preliminary,
  title={Preliminary analysis of the impact of lab results on large language model generated differential diagnoses},
  author={Bhasuran, Balu and Jin, Qiao and Xie, Yuzhang and Yang, Carl and Hanna, Karim and Costa, Jennifer and Shavor, Cindy and Han, Wenshan and Lu, Zhiyong and He, Zhe},
  journal={npj Digital Medicine},
  volume={8},
  number={1},
  pages={166},
  year={2025},
  publisher={Nature Publishing Group UK London}
}

@article{us2024transparency,
  title={Transparency for machine learning-enabled medical devices: Guiding principles},
  author={US Food and Drug Administration and others},
  journal={US Food And Drug Administration. Retrieved June},
  volume={30},
  pages={2024},
  year={2024}
}

@article{babic2025general,
  title={A general framework for governing marketed AI/ML medical devices},
  author={Babic, Boris and Glenn Cohen, I and Stern, Ariel Dora and Li, Yiwen and Ouellet, Melissa},
  journal={npj Digital Medicine},
  volume={8},
  number={1},
  pages={328},
  year={2025},
  publisher={Nature Publishing Group UK London}
}

@article{kore2024empirical,
  title={Empirical data drift detection experiments on real-world medical imaging data},
  author={Kore, Ali and Abbasi Bavil, Elyar and Subasri, Vallijah and Abdalla, Moustafa and Fine, Benjamin and Dolatabadi, Elham and Abdalla, Mohamed},
  journal={Nature communications},
  volume={15},
  number={1},
  pages={1887},
  year={2024},
  publisher={Nature Publishing Group UK London}
}

@article{subasri2025detecting,
  title={Detecting and Remediating Harmful Data Shifts for the Responsible Deployment of Clinical AI Models},
  author={Subasri, Vallijah and Krishnan, Amrit and Kore, Ali and Dhalla, Azra and Pandya, Deval and Wang, Bo and Malkin, David and Razak, Fahad and Verma, Amol A and Goldenberg, Anna and others},
  journal={JAMA Network Open},
  volume={8},
  number={6},
  pages={e2513685--e2513685},
  year={2025},
  publisher={American Medical Association}
}

@article{charlin2007scripts,
  title={Scripts and clinical reasoning},
  author={Charlin, Bernard and Boshuizen, Henny PA and Custers, Eugene J and Feltovich, Paul J},
  journal={Medical education},
  volume={41},
  number={12},
  pages={1178--1184},
  year={2007},
  publisher={Wiley Online Library}
}

@article{schwartzstein2025critical,
  title={Critical Thinking for 21st-Century Medicine—Moving Beyond Illness Scripts},
  author={Schwartzstein, Richard M and Iyer, Alexander A},
  journal={JAMA},
  volume={334},
  number={17},
  pages={1509--1510},
  year={2025},
  publisher={American Medical Association}
}

@article{singh2015advancing,
  title={Advancing the science of measurement of diagnostic errors in healthcare: the Safer Dx framework},
  author={Singh, Hardeep and Sittig, Dean F},
  journal={BMJ quality \& safety},
  volume={24},
  number={2},
  pages={103--110},
  year={2015},
  publisher={BMJ Publishing Group Ltd}
}

@article{dalal2025adverse,
  title={Adverse diagnostic events in hospitalised patients: a single-centre, retrospective cohort study},
  author={Dalal, Anuj K and Plombon, Savanna and Konieczny, Kaitlyn and Motta-Calderon, Daniel and Malik, Maria and Garber, Alison and Lam, Alyssa and Piniella, Nicholas and Leeson, Marie and Garabedian, Pamela and others},
  journal={BMJ Quality \& Safety},
  volume={34},
  number={6},
  pages={377--388},
  year={2025},
  publisher={BMJ Publishing Group Ltd}
}

@article{nenadic_physicians_2026,
	title = {Physicians as context engineers in the era of generative {AI}},
	issn = {1546-170X},
	url = {https://doi.org/10.1038/s41591-026-04215-x},
	doi = {10.1038/s41591-026-04215-x},
	journal = {Nature Medicine},
	author = {Nenadic, Ivan and Fudim, Marat and Loring, Zak and Sharma, Abhinav and Fedacko, Jan and Murthy, Venkatesh and Piccini, Jonathan},
	month = feb,
	year = {2026},
}

@article{mahajan2025cognitive,
  title={Cognitive bias in clinical large language models},
  author={Mahajan, Arjun and Obermeyer, Ziad and Daneshjou, Roxana and Lester, Jenna and Powell, Dylan},
  journal={npj Digital Medicine},
  volume={8},
  number={1},
  pages={428},
  year={2025},
  publisher={Nature Publishing Group UK London}
}

@misc{bge_embedding,
      title={C-Pack: Packaged Resources To Advance General Chinese Embedding}, 
      author={Shitao Xiao and Zheng Liu and Peitian Zhang and Niklas Muennighoff},
      year={2023},
      eprint={2309.07597},
      archivePrefix={arXiv},
      primaryClass={cs.CL}
}

@article{yang2025qwen3,
  title={Qwen3 technical report},
  author={Yang, An and Li, Anfeng and Yang, Baosong and Zhang, Beichen and Hui, Binyuan and Zheng, Bo and Yu, Bowen and Gao, Chang and Huang, Chengen and Lv, Chenxu and others},
  journal={arXiv preprint arXiv:2505.09388},
  year={2025}
}

@article{liu2025deepseek,
  title={Deepseek-v3. 2: Pushing the frontier of open large language models},
  author={Liu, Aixin and Mei, Aoxue and Lin, Bangcai and Xue, Bing and Wang, Bingxuan and Xu, Bingzheng and Wu, Bochao and Zhang, Bowei and Lin, Chaofan and Dong, Chen and others},
  journal={arXiv preprint arXiv:2512.02556},
  year={2025}
}

@misc{5team2025glm45agenticreasoningcoding,
      title={GLM-4.5: Agentic, Reasoning, and Coding (ARC) Foundation Models}, 
      author={GLM Team and Aohan Zeng and Xin Lv and Qinkai Zheng and Zhenyu Hou and Bin Chen and Chengxing Xie and Cunxiang Wang and Da Yin and Hao Zeng and Jiajie Zhang and Kedong Wang and Lucen Zhong and Mingdao Liu and Rui Lu and Shulin Cao and Xiaohan Zhang and Xuancheng Huang and Yao Wei and Yean Cheng and Yifan An and Yilin Niu and Yuanhao Wen and Yushi Bai and Zhengxiao Du and Zihan Wang and Zilin Zhu and Bohan Zhang and Bosi Wen and Bowen Wu and Bowen Xu and Can Huang and Casey Zhao and Changpeng Cai and Chao Yu and Chen Li and Chendi Ge and Chenghua Huang and Chenhui Zhang and Chenxi Xu and Chenzheng Zhu and Chuang Li and Congfeng Yin and Daoyan Lin and Dayong Yang and Dazhi Jiang and Ding Ai and Erle Zhu and Fei Wang and Gengzheng Pan and Guo Wang and Hailong Sun and Haitao Li and Haiyang Li and Haiyi Hu and Hanyu Zhang and Hao Peng and Hao Tai and Haoke Zhang and Haoran Wang and Haoyu Yang and He Liu and He Zhao and Hongwei Liu and Hongxi Yan and Huan Liu and Huilong Chen and Ji Li and Jiajing Zhao and Jiamin Ren and Jian Jiao and Jiani Zhao and Jianyang Yan and Jiaqi Wang and Jiayi Gui and Jiayue Zhao and Jie Liu and Jijie Li and Jing Li and Jing Lu and Jingsen Wang and Jingwei Yuan and Jingxuan Li and Jingzhao Du and Jinhua Du and Jinxin Liu and Junkai Zhi and Junli Gao and Ke Wang and Lekang Yang and Liang Xu and Lin Fan and Lindong Wu and Lintao Ding and Lu Wang and Man Zhang and Minghao Li and Minghuan Xu and Mingming Zhao and Mingshu Zhai and Pengfan Du and Qian Dong and Shangde Lei and Shangqing Tu and Shangtong Yang and Shaoyou Lu and Shijie Li and Shuang Li and Shuang-Li and Shuxun Yang and Sibo Yi and Tianshu Yu and Wei Tian and Weihan Wang and Wenbo Yu and Weng Lam Tam and Wenjie Liang and Wentao Liu and Xiao Wang and Xiaohan Jia and Xiaotao Gu and Xiaoying Ling and Xin Wang and Xing Fan and Xingru Pan and Xinyuan Zhang and Xinze Zhang and Xiuqing Fu and Xunkai Zhang and Yabo Xu and Yandong Wu and Yida Lu and Yidong Wang and Yilin Zhou and Yiming Pan and Ying Zhang and Yingli Wang and Yingru Li and Yinpei Su and Yipeng Geng and Yitong Zhu and Yongkun Yang and Yuhang Li and Yuhao Wu and Yujiang Li and Yunan Liu and Yunqing Wang and Yuntao Li and Yuxuan Zhang and Zezhen Liu and Zhen Yang and Zhengda Zhou and Zhongpei Qiao and Zhuoer Feng and Zhuorui Liu and Zichen Zhang and Zihan Wang and Zijun Yao and Zikang Wang and Ziqiang Liu and Ziwei Chai and Zixuan Li and Zuodong Zhao and Wenguang Chen and Jidong Zhai and Bin Xu and Minlie Huang and Hongning Wang and Juanzi Li and Yuxiao Dong and Jie Tang},
      year={2025},
      eprint={2508.06471},
      archivePrefix={arXiv},
      url={https://arxiv.org/abs/2508.06471}, 
}

@article{sellergren2025medgemma,
    title = {MedGemma Technical Report},
    author = {Sellergren, Andrew and Kazemzadeh, Sahar and Jaroensri, Tiam and Kiraly, Atilla and Traverse, Madeleine and Kohlberger, Timo and Xu, Shawn and Jamil, Fayaz and Hughes, Cían and Lau, Charles and others},
    journal={arXiv preprint arXiv:2507.05201},
    year = {2025},
}

@article{toma2023clinical,
  title={Clinical camel: An open expert-level medical language model with dialogue-based knowledge encoding},
  author={Toma, Augustin and Lawler, Patrick R and Ba, Jimmy and Krishnan, Rahul G and Rubin, Barry B and Wang, Bo},
  journal={arXiv preprint arXiv:2305.12031},
  year={2023}
}

@article{bean_reliability_2026,
	title = {Reliability of {LLMs} as medical assistants for the general public: a randomized preregistered study},
	issn = {1546-170X},
	url = {https://doi.org/10.1038/s41591-025-04074-y},
	doi = {10.1038/s41591-025-04074-y},
	journal = {Nature Medicine},
	author = {Bean, Andrew M. and Payne, Rebecca Elizabeth and Parsons, Guy and Kirk, Hannah Rose and Ciro, Juan and Mosquera-Gómez, Rafael and Hincapié M, Sara and Ekanayaka, Aruna S. and Tarassenko, Lionel and Rocher, Luc and Mahdi, Adam},
	month = feb,
	year = {2026},
}

@article{ferber2025development,
  title={Development and validation of an autonomous artificial intelligence agent for clinical decision-making in oncology},
  author={Ferber, Dyke and El Nahhas, Omar SM and W{\"o}lflein, Georg and Wiest, Isabella C and Clusmann, Jan and Le{\ss}mann, Marie-Elisabeth and Foersch, Sebastian and Lammert, Jacqueline and Tschochohei, Maximilian and J{\"a}ger, Dirk and others},
  journal={Nature cancer},
  pages={1--13},
  year={2025},
  publisher={Nature Publishing Group US New York}
}

@article{johnson2019billion,
  title={Billion-scale similarity search with {GPUs}},
  author={Johnson, Jeff and Douze, Matthijs and J{\'e}gou, Herv{\'e}},
  journal={IEEE Transactions on Big Data},
  volume={7},
  number={3},
  pages={535--547},
  year={2019},
  publisher={IEEE}
}

@article{moor2023foundation,
  title={Foundation models for generalist medical artificial intelligence},
  author={Moor, Michael and Banerjee, Oishi and Abad, Zahra Shakeri Hossein and Krumholz, Harlan M and Leskovec, Jure and Topol, Eric J and Rajpurkar, Pranav},
  journal={Nature},
  volume={616},
  number={7956},
  pages={259--265},
  year={2023},
  publisher={Nature Publishing Group UK London}
}

@article{rajpurkar2022ai,
  title={AI in health and medicine},
  author={Rajpurkar, Pranav and Chen, Emma and Banerjee, Oishi and Topol, Eric J},
  journal={Nature medicine},
  volume={28},
  number={1},
  pages={31--38},
  year={2022},
  publisher={Nature Publishing Group US New York}
}

@article{di2020diagnosis,
  title={Diagnosis and treatment of acute appendicitis: 2020 update of the WSES Jerusalem guidelines},
  author={Di Saverio, Salomone and Podda, Mauro and De Simone, Belinda and Ceresoli, Marco and Augustin, Goran and Gori, Alice and Boermeester, Marja and Sartelli, Massimo and Coccolini, Federico and Tarasconi, Antonio and others},
  journal={World journal of emergency surgery},
  volume={15},
  number={1},
  pages={27},
  year={2020},
  publisher={Springer}
}

@article{yokoe2018tokyo,
  title={Tokyo Guidelines 2018: diagnostic criteria and severity grading of acute cholecystitis (with videos)},
  author={Yokoe, Masamichi and Hata, Jiro and Takada, Tadahiro and Strasberg, Steven M and Asbun, Horacio J and Wakabayashi, Go and Kozaka, Kazuto and Endo, Itaru and Deziel, Daniel J and Miura, Fumihiko and others},
  journal={Journal of Hepato-biliary-pancreatic Sciences},
  volume={25},
  number={1},
  pages={41--54},
  year={2018},
  publisher={Wiley Online Library}
}

@article{sartelli20202020,
  title={2020 update of the WSES guidelines for the management of acute colonic diverticulitis in the emergency setting},
  author={Sartelli, Massimo and Weber, Dieter G and Kluger, Yoram and Ansaloni, Luca and Coccolini, Federico and Abu-Zidan, Fikri and Augustin, Goran and Ben-Ishay, Offir and Biffl, Walter L and Bouliaris, Konstantinos and others},
  journal={World Journal of Emergency Surgery},
  volume={15},
  number={1},
  pages={32},
  year={2020},
  publisher={Springer}
}

@article{leppaniemi20192019,
  title={2019 WSES guidelines for the management of severe acute pancreatitis},
  author={Lepp{\"a}niemi, Ari and Tolonen, Matti and Tarasconi, Antonio and Segovia-Lohse, Helmut and Gamberini, Emiliano and Kirkpatrick, Andrew W and Ball, Chad G and Parry, Neil and Sartelli, Massimo and Wolbrink, Daan and others},
  journal={World journal of emergency surgery},
  volume={14},
  number={1},
  pages={27},
  year={2019},
  publisher={Springer}
}

@article{li2024agent,
  title={Agent hospital: A simulacrum of hospital with evolvable medical agents},
  author={Li, Junkai and Lai, Yunghwei and Li, Weitao and Ren, Jingyi and Zhang, Meng and Kang, Xinhui and Wang, Siyu and Li, Peng and Zhang, Ya-Qin and Ma, Weizhi and others},
  journal={arXiv preprint arXiv:2405.02957},
  year={2024}
}

@article{tu2025towards,
  title={Towards conversational diagnostic artificial intelligence},
  author={Tu, Tao and Schaekermann, Mike and Palepu, Anil and Saab, Khaled and Freyberg, Jan and Tanno, Ryutaro and Wang, Amy and Li, Brenna and Amin, Mohamed and Cheng, Yong and others},
  journal={Nature},
  volume={642},
  number={8067},
  pages={442--450},
  year={2025},
  publisher={Nature Publishing Group UK London}
}

@article{chen2025enhancing,
  title={Enhancing diagnostic capability with multi-agents conversational large language models},
  author={Chen, Xi and Yi, Huahui and You, Mingke and Liu, WeiZhi and Wang, Li and Li, Hairui and Zhang, Xue and Guo, Yingman and Fan, Lei and Chen, Gang and others},
  journal={NPJ digital medicine},
  volume={8},
  number={1},
  pages={159},
  year={2025},
  publisher={Nature Publishing Group UK London}
}

@article{liu2025generalist,
  title={A generalist medical language model for disease diagnosis assistance},
  author={Liu, Xiaohong and Liu, Hao and Yang, Guoxing and Jiang, Zeyu and Cui, Shuguang and Zhang, Zhaoze and Wang, Huan and Tao, Liyuan and Sun, Yongchang and Song, Zhu and others},
  journal={Nature medicine},
  volume={31},
  number={3},
  pages={932--942},
  year={2025},
  publisher={Nature Publishing Group US New York}
}

@inproceedings{xu2025amem,
  title={A-mem: Agentic memory for llm agents},
  editor="Zaimis, E.",
  author={Xu, Wujiang and Liang, Zujie and Mei, Kai and Gao, Hang and Tan, Juntao and Zhang, Yongfeng},
  booktitle={Advances in Neural Information Processing Systems},
  year={2025}
}

@inproceedings{dong2025memory,
  title={Memory Injection Attacks on LLM Agents via Query-Only Interaction},
  editor="Zaimis, E.",
  author={Dong, Shen and Xu, Shaochen and He, Pengfei and Li, Yige and Tang, Jiliang and Liu, Tianming and Liu, Hui and Xiang, Zhen},
  booktitle={Advances in Neural Information Processing Systems},
  year={2025}
}

@article{nori2025sequential,
  title={Sequential diagnosis with language models},
  author={Nori, Harsha and Daswani, Mayank and Kelly, Christopher and Lundberg, Scott and Ribeiro, Marco Tulio and Wilson, Marc and Liu, Xiaoxuan and Sounderajah, Viknesh and Carlson, Jonathan and Lungren, Matthew P and others},
  journal={arXiv preprint arXiv:2506.22405},
  year={2025}
}

@article{zhao2026agentic,
  title={An agentic system for rare disease diagnosis with traceable reasoning},
  author={Zhao, Weike and Wu, Chaoyi and Fan, Yanjie and Qiu, Pengcheng and Zhang, Xiaoman and Sun, Yuze and Zhou, Xiao and Zhang, Shuju and Peng, Yu and Wang, Yanfeng and others},
  journal={Nature},
  pages={1--10},
  year={2026},
  publisher={Nature Publishing Group UK London}
}
